\documentclass{article}
\usepackage{iclr2027_conference,times}

\usepackage{amsmath,amsfonts,bm}

\def\eqref#1{equation~\ref{#1}}

\def\1{\bm{1}}

\DeclareMathAlphabet{\mathsfit}{\encodingdefault}{\sfdefault}{m}{sl}
\SetMathAlphabet{\mathsfit}{bold}{\encodingdefault}{\sfdefault}{bx}{n}

\usepackage{hyperref}
\usepackage{url}

\usepackage{graphicx}
\usepackage{xcolor}
\usepackage{booktabs}
\usepackage{array}
\usepackage{tabularx}
\usepackage{makecell}
\usepackage{multirow}
\usepackage{adjustbox}
\usepackage[textsize=scriptsize]{todonotes}
\usepackage{subcaption}
\usepackage{stfloats}
\usepackage{cleveref}
\usepackage{enumitem}
\usepackage{graphicx}
\usepackage{tcolorbox}
\usepackage{amsthm}
\usepackage{mathtools}
\usepackage{xcolor}
\usepackage{xspace}
\usepackage{bbm}
\usepackage{soul}
\usepackage{color}
\newtheorem{definition}{Definition}

\usepackage{wrapfig}
\newcommand{\fondueicon}{%
  \raisebox{-0.25\height}{%
    \includegraphics[height=1.5em]{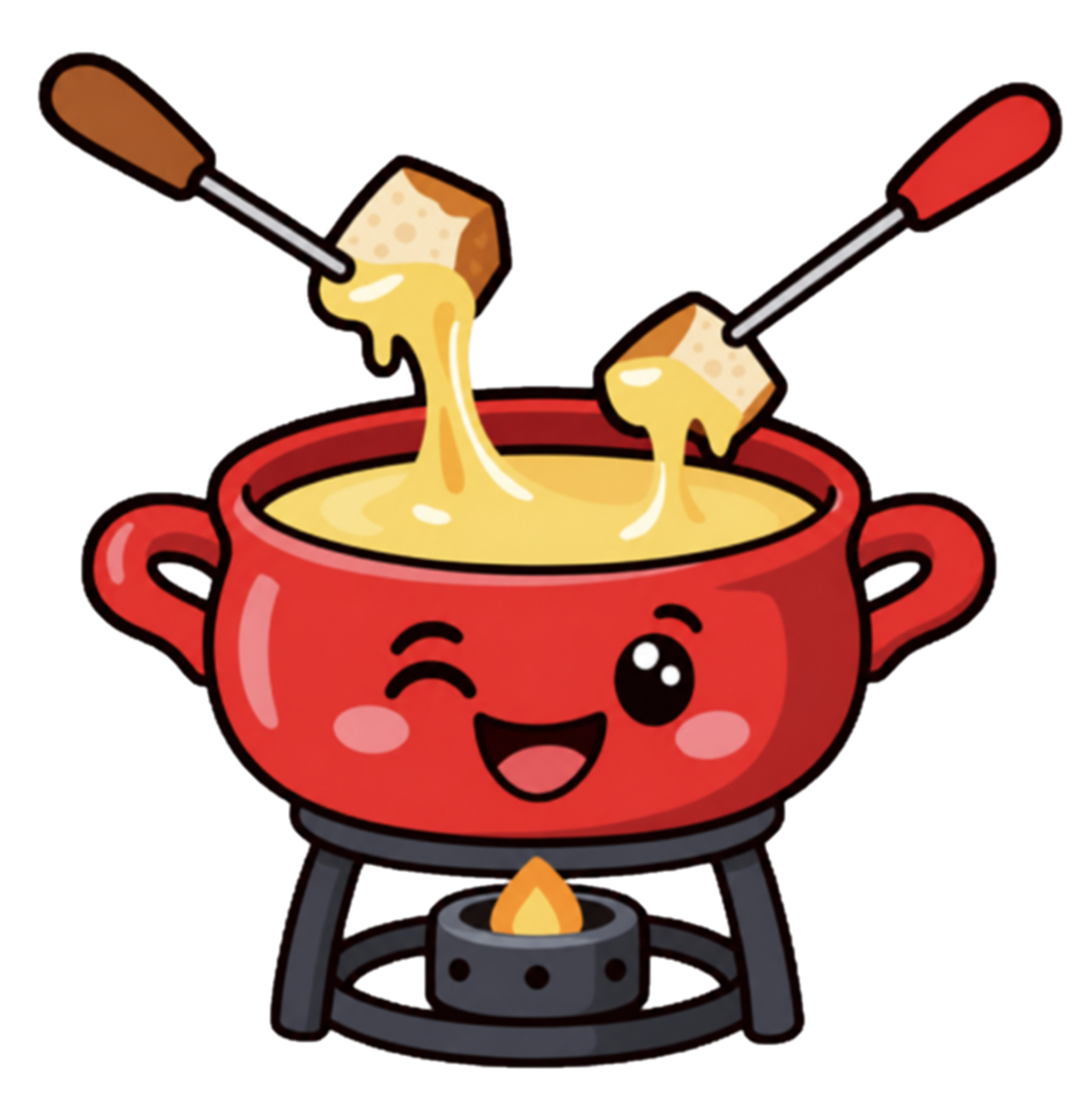}%
  }%
}

\definecolor{bridge}{HTML}{0072B2}
\definecolor{target}{HTML}{E69F00}
\definecolor{competitor}{HTML}{CC3311}

\newcommand{\tmpbridgeonly}{\textcolor{bridge}{Bridge-only}\xspace}
\newcommand{\tmptargetonly}{\textcolor{target}{Target-only}\xspace}

\definecolor{bridge}{HTML}{0072B2}
\definecolor{target}{HTML}{E69F00}
\definecolor{competitor}{HTML}{CC3311}

\newcommand{\bridgefact}{\textcolor{bridge}{bridge fact}\xspace}
\newcommand{\targetfact}{\textcolor{target}{target fact}\xspace}

\usepackage{tikz}
\usetikzlibrary{arrows.meta, positioning, shapes.geometric}
\usepackage{pgfplots}
\pgfplotsset{compat=1.18}

\usepackage{pifont}
\usepackage{tikz}
\usepackage[table]{xcolor}

\newcommand{\cmark}{\ding{51}}
\newcommand{\xmark}{\ding{55}}

\newcommand{\benchmark}{FondueQA}

\newcolumntype{Y}{>{\raggedright\arraybackslash}X}
\newcolumntype{C}[1]{>{\centering\arraybackslash}p{#1}}
\newcolumntype{L}[1]{>{\raggedright\arraybackslash}p{#1}}
\newcolumntype{R}[1]{>{\raggedleft\arraybackslash}p{#1}}

\newcommand{\fondueTableSetup}{%
  \small
  \setlength{\tabcolsep}{4pt}%
  \renewcommand{\arraystretch}{1.12}%
}

\newcommand{\hotpot}{\textsc{HotpotQA}}
\newcommand{\twowiki}{\textsc{2WikiMultiHopQA}}
\newcommand{\popqa}{\textsc{PopQA}}
\newcommand{\mintqa}{\textsc{MINTQA}}
\newcommand{\musique}{\textsc{MuSiQue-Ans}}
\newcommand{\fondue}{\textsc{FondueQA}}
\newcommand{\hotpotext}{\textsc{HotpotQA-Ext}}
\newcommand{\twowikiext}{\textsc{2WikiMultiHopQA-Ext}}

\newcommand{\mintqaext}{\textsc{MINTQA-Ext}}
\newcommand{\musiqueext}{\textsc{MuSiQue-Ext}}
\newcommand{\ext}{\textsc{-Ext}}

\definecolor{bridgec}{RGB}{0,114,178}       
\definecolor{entryc}{RGB}{0,158,115}        
\definecolor{entityonec}{RGB}{204,121,167}  
\definecolor{entitytwoc}{RGB}{230,159,0}   
\definecolor{targetc}{RGB}{213,94,0}       

\definecolor{basebar}{RGB}{76,120,168}      
\definecolor{gainbar}{RGB}{245,133,24}    

\definecolor{fdDark}{RGB}{31,64,104}    
\definecolor{fdMed}{RGB}{74,123,167}     
\definecolor{fdLight}{RGB}{162,197,222}  
\definecolor{fdAccent}{RGB}{196,90,44}   
\definecolor{fdGrid}{RGB}{223,227,232} 
\definecolor{fdBorder}{RGB}{40,40,45}

\pgfplotsset{
  fonduebar/.style={
    ybar,
    width=0.90\linewidth,
    height=4.6cm,
    ymin=0,
    axis x line*=bottom,
    axis y line*=left,
    tick align=outside,
    major grid style={fdGrid},
    ymajorgrids=true,
    tick label style={font=\footnotesize},
    label style={font=\small},
    legend style={
      draw=none,
      fill=none,
      font=\footnotesize,
      at={(0.5,1.05)},
      anchor=south,
      legend columns=-1
    },
    clip=false
  },
  fonduehbar/.style={
    xbar,
    width=0.92\linewidth,
    height=5.8cm,
    xmin=0,
    axis x line*=bottom,
    axis y line*=left,
    tick align=outside,
    major grid style={fdGrid},
    xmajorgrids=true,
    tick label style={font=\footnotesize},
    label style={font=\small},
    legend style={
      draw=none,
      fill=none,
      font=\footnotesize,
      at={(0.5,1.05)},
      anchor=south,
      legend columns=-1
    },
    clip=false
  }
}

\usepackage{geometry}
\title{\fondueicon~FondueQA: A Benchmark for Evaluating Parametric Compensation in Multi-Hop QA}

\author{Laurin Schmid, Suman Navaratnarajah, Koel Dutta Chowdhury,\\
\textbf{Steffen Eger, Josif Grabocka, Christian Frey} \\
Department of Computer Science \& Artificial Intelligence\\
University of Technology Nuremberg\\
Dr.-Luise-Herzberg-Straße 4, 90461 Nuremberg, Germany\\
\texttt{\{firstname.lastname\}@utn.de}
}

\usepackage[dvipsnames]{xcolor}
\usepackage{newunicodechar}
\usepackage{soul}
\newunicodechar{♫}{}
\newunicodechar{♬}{}
\newunicodechar{♪}{}

\iclrfinalcopy

\begin{document}

\maketitle
\fancyhead[L]{Preprint}
\pagestyle{fancy}
\thispagestyle{fancy}
\begin{abstract}  
Large Language Models (LLMs) show remarkable performance based on parametric knowledge, i.e., their ability to recall facts and associations internalized in a model's weights. However, studying this capability on established multi-hop benchmarks may be complicated by benchmark contamination, as many of these benchmarks have been publicly available long enough to plausibly appear in modern pretraining corpora. As a result, correct answers under missing evidence may reflect either general parametric knowledge or memorization of benchmark-specific instances. To study how models use internal knowledge when retrieval evidence is incomplete, we introduce \textbf{\fondue}, in which every required multi-hop fact is uniquely localized to a designated evidence document. This enables controlled document-level evidence ablations to study parametric compensation under incomplete retrieval. Our empirical evaluation reveals that parametric compensation strongly depends on both entity popularity and the structural role of missing evidence. Generally, models compensate more successfully for facts involving popular entities. On \textbf{\fondue}, we show that bridging evidence can be more useful than the answer-bearing document itself. Moreover, models may recall a missing fact in isolation but fail to use it under partial evidence, particularly when the context contains plausible competing entities, which can reduce Exact Match by up to 15 percentage points.
\end{abstract}
\section{Introduction}
\label{sec:introduction}

Large Language Models (LLMs) are pre-trained on a large corpus of factual knowledge that is stored within their weights, referred to as parametric knowledge \citep{Zhao_2026_SurveyLLM}. Whenever user queries and requests go beyond the internalized knowledge, LLMs are increasingly paired with retrieval systems that provide external evidence at inference time, improving factual accuracy \citep{Ni_2026_TrustworthyRAG,gao_2024_RAGLLM,Arslan_2024_SurveyRAG}. Entity popularity is associated with differences in factual QA performance and the gains obtained from retrieval augmentation \citep{Yang_2024_CRAG}. Complementary theoretical work models the benefits and risks of retrieved information at the token level \citep{Xu_2025_TokenLevelHarmonization}. At the same time, LLMs can sometimes answer correctly even when the retrieved context is insufficient to support the answer \citep{joren2025sufficient}. However, in multi-hop QA, the utility of a retrieved passage depends on its relationship to the other passages in the context \citep{jain-garimella-2025-modeling}. 

Prior work studies this interaction for single-fact questions, showing that models recall popular facts more reliably and benefit more from retrieval for less popular information \citep{mallen-etal-2023-trust,Yang_2024_CRAG}. Multi-hop questions add structure to this interaction. Consider the query \textit{``What is the nationality of the director of the film Titanic?''}. Answering it requires knowing the interdependent facts that \emph{Titanic} was directed by \textit{James Cameron}, and that \textit{James Cameron} is \textit{Canadian}. With only part of this evidence observed, the question becomes whether the model can recover the missing link, either from \emph{Titanic} to \emph{James Cameron} or from \emph{James Cameron} to his nationality. How evidence structure affects compensation, and which other factors determine success under partial retrieval, remain largely unexplored.

We argue that existing multi-hop benchmarks provide important components for this analysis, but were not designed for a controlled removal of individual reasoning steps (\Cref{tab:benchmark_comparison}). While \popqa\ \citep{mallen-etal-2023-trust} provides entity-popularity measures and \mintqa\ \citep{he-etal-2026-mintqa} extends this to multi-hop, they do not provide context documents. Further, \twowiki\ and \musique\ \citep{ho-etal-2020-constructing,trivedi-etal-2022-musique} move towards stronger multi-hop supervision, but focus on reasoning paths or connected reasoning rather than explicit fact isolation required for controlled fact removal. Consequently, we observe (i) \textit{Question leakage}, where the question reveals (parts of) the answer; (ii) \textit{Cross-document leakage}, where another document reveals a questioned fact; (iii) \textit{Distractor leakage}, where an allegedly irrelevant document contains required evidence. 

To enable a controllable investigation over facts, we propose a novel multi-hop QA benchmark, called \fondue. In a first step, we leverage a knowledge graph as the source of candidate facts. In a second step, we construct template-based multi-hop questions from these facts. In the third step, our pipeline ensures that each fact is assigned to exclusively one gold document, which we refer to as \emph{Fact Isolation}. Subsequently, we conduct a thorough validation and filtering step, where it is ensured that the required facts are supported within their assigned gold document without any cross leakage of required evidence. Rather than altering the original documents, we discard invalid instances unless the generated question-answer pair can be adjusted without changing the intended evidence structure. As a result, each document-level ablation induces a controlled removal of its assigned fact, enabling direct comparisons across partial evidence conditions.

In our empirical study, we show that parametric compensation under partial evidence depends not only on the accessibility of missing facts, but also on their structural role in the reasoning path. On \fondue, bridging evidence (e.g. 'James Cameron directed Titanic') can be more valuable than the answer-bearing document itself (e.g. 'James Cameron is Canadian'), while facts that are successfully recalled in isolation may still fail to contribute to the answer. Compensation is further shaped by the surrounding context, as plausible competing entities can reduce \textit{Exact Match} by up to 15 percentage points when supporting alternative answers, but may improve it by more than 25 points when supporting the correct answer.

We make three main contributions: 
\begin{itemize}[leftmargin=*]
     \item \textbf{Controlled benchmark construction.} We introduce a benchmark construction pipeline that enforces fact isolation and controls for evidence leakage. Using this pipeline, we construct \fondue, covering different question types, evidence structures, answer formats, and role-specific popularity information. We also adapt established multi-hop benchmarks for the same controlled analysis.
      \item \textbf{Determinants of parametric compensation.} We show that compensation for missing evidence depends on entity popularity and evidence role. On \fondue, bridge evidence can be more useful than the document containing the target although the utility of evidence roles varies across benchmarks.
    \item \textbf{Limits of parametric compensation.} We show that recalling a missing fact in isolation does not guarantee its use under partial evidence, and that plausible competing entities can substantially change whether parametric knowledge is successfully integrated into the reasoning path.
\end{itemize}

\section{Related Work}
\label{sec:related_work}

\textbf{Benchmarks.~}The original \hotpot\ \citep{yang-etal-2018-hotpotqa} introduced Wikipedia-based QA pairs to benchmark multi-document reasoning. \twowiki\ \citep{ho-etal-2020-constructing} provides a complete reasoning path for the multi-hop questions aimed at measuring the true reasoning skill of models rather than relying on shortcuts. \popqa\ \citep{mallen-etal-2023-trust} studies single factual question answering across entities with varying popularity, highlighting the difficulty of recalling unpopular knowledge from model parameters. \mintqa\ \citep{he-etal-2026-mintqa} extends this perspective to multi-hop questions with varying levels of knowledge freshness and entity popularity. \musique\ \citep{trivedi-etal-2022-musique} was designed to make it harder to answer through disconnected reasoning. \textsc{MuSiQue-Full} \citep{trivedi-etal-2022-musique} also constructs unanswerable variants by removing support for a selected subquestion. Our construction instead targets a unique fact-to-document assignment for controlled, role-specific evidence ablations. Taken together, existing benchmarks provide complementary components for our analysis (\Cref{tab:benchmark_comparison}). \popqa\ and \mintqa\ provide entity popularity information, while several multi-hop benchmarks provide context documents or explicit subquestions. Some benchmarks additionally provide answer aliases for a more robust evaluation. \fondue\ further follows a versioned live-benchmark setup in which new instances will be released over time.

\textbf{Parametric compensation.~}Prior work investigates the effect of entity popularity and parametric compensation with a focus on improving retrieval augmentation, however they are constrained to single-hop questions \citep{mallen-etal-2023-trust}. \citet{joren2025sufficient} demonstrate that LLMs can answer correctly even when the retrieved context is insufficient, highlighting that external evidence might not need to contain all information required for an answer. Other studies further investigate how models balance parametric and contextual knowledge in retrieval-augmented generation \citep{farahani-johansson-2024-deciphering,bi_2026_parameters}. Related work on answerability shows that recognizing unanswerable questions becomes harder when the context contains plausible same-type candidates \citep{sulem-etal-2021-know-dont}. Whether such competition affects parametric compensation under controlled multi-hop evidence gaps remains unexplored.

\textbf{Evaluation Bias.~}Benchmark contamination occurs when evaluation items or information derived from them are included in model training, potentially inflating measured performance \citep{xu_2024_benchmarkdatacontaminationlarge}. \citet{dong_2024_generalizationMemorization} inspect potential contamination from the concentration of sampled model outputs. \citet{deng_2024_dataContamination} show that commercial LLMs can reconstruct masked incorrect answer options and unlikely words from benchmark instances, providing evidence of potential benchmark contamination. Earlier studies show that high performance on multi-hop QA benchmarks can arise from dataset artifacts and disconnected reasoning rather than the intended composition of supporting facts \citep{guo-etal-2023-counterfactual,chen-durrett-2019-understanding, trivedi-etal-2020-multihop}. More recently, \citet{fang-etal-2025-lastingbench} identify behavior consistent with knowledge leakage through context-removal and question-perturbation tests.

\begin{table*}[t]
\caption{
\textbf{Comparison of FondueQA to existing QA benchmarks.}
}
\centering

\setlength{\tabcolsep}{12pt}

\begin{adjustbox}{max width=\textwidth}
\begin{tabular}{l c c c c c c c}
\toprule
& & & & & & \multicolumn{2}{c}{\textbf{Evidence Control}} \\
\cmidrule(lr){7-8}

\textbf{Benchmark}
  & \textbf{Live Bench}
  & \textbf{Answer Aliases}
  & \textbf{Popularity}
  & \textbf{Context Documents}
  & \textbf{Subquestions}
  & \makecell{\textbf{Distractor}\\\textbf{Checks}}
  & \makecell{\textbf{Fact}\\\textbf{Isolation}}
\\
\midrule

HotpotQA
  & \xmark
  & \xmark
  & \xmark
  & \cmark
  & \xmark
  & \xmark
  & \xmark
\\

MINTQA
  & \xmark
  & \xmark
  & \cmark
  & \xmark
  & \cmark
  & --
  & --
\\

2WikiMultiHopQA
  & \xmark
  & \cmark
  & \xmark
  & \cmark
  & \xmark
  & \xmark
  & \xmark
\\

PopQA
  & \xmark
  & \cmark
  & \cmark
  & \xmark
  & --
  & --
  & --
\\

MuSiQue-Ans
  & \xmark
  & \cmark
  & \xmark
  & \cmark
  & \cmark
  & \xmark
  & \xmark
\\

\midrule
\rowcolor{blue!6}
\textbf{FondueQA}
  & \textbf{\cmark}
  & \textbf{\cmark}
  & \textbf{\cmark}
  & \textbf{\cmark}
  & \textbf{\cmark}
  & \textbf{\cmark}
  & \textbf{\cmark}
\\

\bottomrule
\end{tabular}
\end{adjustbox}

\label{tab:benchmark_comparison}
\end{table*}
\section{Benchmark Construction Pipeline}

\label{sec:method}
We construct \fondue\ to support controlled removal of individual facts from multi-hop evidence. Our pipeline generates multi-hop questions from knowledge graph triples and assigns each required fact to one document. It then verifies the fact is absent from the question and all other documents, including distractors. Invalid instances are repaired or discarded. Thus, removing a gold document removes its assigned fact. 

Our construction supports two interventions. 
An \emph{evidence gap} removes one or more documents and, consequently, the required facts from the context. A \emph{reasoning trap} keeps this gap fixed but adds plausible competing entities to the context. We formalize both interventions below and describe the resulting benchmark.

\subsection{Formal Setting}
\label{subsec:formalSetting}

Let $\mathcal{KG}=(\mathcal{E},\mathcal{R},\mathcal{T})$ be a \textit{Knowledge Graph} with entities $\mathcal{E}$, relations $\mathcal{R}$, and facts (triples) $\mathcal{T}\subseteq\mathcal{E}\times\mathcal{R}\times\mathcal{E}$. We use Wikidata as $\mathcal{KG}$ and link each entity $e$ to its Wikipedia article $\operatorname{doc}(e)$, which serves as the document that expresses facts about that entity. We assume a closed world, i.e., a triple that is not in $\mathcal{KG}$ is treated as unknown.
We consider a fact $f=(s,r,o)$ is expressed by the associated document of its subject, whereas the support of $f$ in the object's document must not hold (cf. \Cref{subsec:requirementAnalysis}). 

A question is constructed from a set of required facts $F=\{f_1,\ldots,f_k\}\subseteq\mathcal{T}.$ The relations among these facts define the question's evidence structure. We refer to the induced subgraph of a selection of required facts as \emph{Evidence Structure}. Given a specific evidence structure we define a reasoning path leveraging required facts:
\begin{definition}[Reasoning Path]
\label{def:reasoningPath}
A reasoning path from $e_0$ to $e_n$ is a sequence of facts $\pi=(f_1,\dots,f_n)$ with $f_i=(e_{i-1},r_i,e_i)\in\mathcal{T}$ and two consecutive facts are linked by a commonly shared entity.
\end{definition}

A QA instance is defined as $x=(q,D,A,F,\Gamma)$: a question $q$, a set of context documents $D\coloneqq D^+\cup D^-$ (gold and distractor documents, $D^+\cap D^-=\emptyset$), a set $A$ of valid answer forms (canonical answer and aliases), the set $F=\{f_1,\dots,f_k\}\subseteq\mathcal{T}$ of facts required to answer $q$, and an assignment $\Gamma:F\to D^+$ with $\Gamma(f)=\operatorname{doc}(s)$ of each required fact $f=(s,r,o)$ to its gold document. For a text $t$ and a fact $f$, let $\operatorname{supp}(t,f)\in\{0,1\}$ indicate whether $f$ can be derived from $t$. Removing one or more required facts from the context results in an \textit{Evidence Gap} defined as:

\begin{definition}[Evidence Gap]
\label{def:evidenceGap}
Let \(G \subseteq F\) be a non-empty set of required facts. The evidence gap induced by \(G\) removes the gold document \(\Gamma(f)\) for every \(f\in G\), yielding the context \(D_{-G}=D\setminus\{\Gamma(f)\mid f\in G\}\). The facts in \(G\) are missing and the facts in \(F\setminus G\) are retained. The gap is \emph{partial} if \(G\neq F\) and \emph{complete} if \(G=F\).
\end{definition}

For this intervention to be controlled, each removed fact must appear only in its assigned document. We state the corresponding requirements in \Cref{subsec:requirementAnalysis}. Furthermore, replacing distractors with plausible competitors leads to a \textit{Reasoning Trap}:

\begin{definition}[Reasoning Trap]
\label{def:reasoningTrap}
Let $f^-{=}(s,r,o)\in G$ be a missing fact of an evidence gap $G\subseteq F$. 
A \emph{competitor} for $f^-$ is a distractor $d\in D^-$ about an alternative entity $o'\neq o$ that is admissible for the same reasoning role and supports no required fact in $F$, i.e., $\operatorname{supp}(d,f)=0$ for all $f\in F$ . A \emph{reasoning trap} replaces distractor documents in $D_{-G}$
with competitors while keeping the missing facts and context size fixed. 
\end{definition}

For example, consider the question \emph{``What is the nationality of the director of \emph{Titanic}?''}. Its reasoning path consists of the \bridgefact $f_1{=}(\text{\emph{Titanic}},\text{director}, \text{James Cameron})$ and the \targetfact $f_2{=}(\text{James Cameron},\text{nationality}, \text{Canada})$. Removing the bridge fact corresponds to the partial evidence gap \(G=\{f_1\}\). The target evidence \(f_2\) remains, while the document identifying James Cameron as the director of \emph{Titanic} is removed. \textcolor{competitor}{Steven Spielberg} or \textcolor{competitor}{Ridley Scott} serve as alternative directors while preserving the gap.

\subsection{Requirements for Controlled Evidence Removal}
\label{subsec:requirementAnalysis}
Existing multi-hop benchmarks can admit reasoning shortcuts that bypass the intended use of supporting documents \citep{jiang-bansal-2019-avoiding,trivedi-etal-2022-musique,trivedi-etal-2020-multihop}. In our validation, we distinguish question leakage, cross-document leakage and required evidence appearing in distractor documents. We impose the following conditions.

\textbf{Fact isolation.}
Each required fact must be supported by its assigned gold document and absent from the remaining context:

\begin{enumerate}[label=(C\arabic*),leftmargin=*,itemsep=1pt,topsep=2pt]
    \item \textbf{Gold support:}
    \(\Gamma\) is injective and
    \(\operatorname{supp}(\Gamma(f),f)=1\) for every \(f\in F\).

    \item \textbf{Isolation from other documents:}
    \(\operatorname{supp}(d,f)=0\) for every \(f\in F\) and every \(d\in D\setminus\{\Gamma(f)\}\).

    \item \textbf{Absence from the question:}
    \(\operatorname{supp}(q,f)=0\) for every \(f\in F\), and \(q\) alone is insufficient.
\end{enumerate}
For required facts \(F=\{f_1,\ldots,f_k\}\) and gold documents
\(D^{+}=\{d_1,\ldots,d_k\}\), define
\[
M_{ij}=\operatorname{supp}(d_j,f_i).
\]
Conditions (C1) and (C2) imply \(M=I_k\) for the gold documents, while (C2) additionally excludes support from distractor documents. Removing \(\Gamma(f_i)\) therefore removes \(f_i\) while leaving the other required facts available.

\textbf{Question identification.}
\label{req:identifiability}
The query $q$ must identify the entities needed to follow the intended evidence structure without revealing a required fact or the answer. Because entity names can be ambiguous \citep{Zhu_2018_Disambiguation, min-etal-2020-ambigqa, lee2024ambigdocs}, we add a disambiguating descriptor when needed for unique reference. The descriptor is retained only if the resulting question continues to satisfy (C3).

\textbf{Evidence structure diversity.}
\label{req:diversity}
The benchmark covers evidence structures (cf. \Cref{subsec:formalSetting}) that differ in fact dependencies, entities exposed in the question, and how the answer is determined. We use manually designed templates for five overarching structures summarized in \Cref{tab:templates}. Additional templates can be added without changing the rest of the pipeline. Sample questions for each type are given in Appendix~\ref{app:templates}.

\begin{table}[t]
\caption{\textbf{Evidence structures.} $v_1,v_2$ denote the property values and $v^\ast$ a fixed reference value.}
\centering
\small
\label{tab:templates}
\begin{tabularx}{\linewidth}{lcX}
\toprule
Structure & $|F|$ & Question $q$ asks for \\
\midrule
Bridge        & 2-3 & Final entity $e_2$ of $(e_0,r_1,e_1),(e_1,r_2,e_2)$ with $e_1$ unnamed in $q$ \\
Comparison    & 2 & Which of two entities has the smaller/larger $v$ (or whether $v_1=v_2$) for the same relation \\
Cross-Property& 2 & Whether $v_1=v_2$ for two different relations \\
Contrastive   & 2 & Which of two entities satisfies $v=v^\ast$: A, B, both, or neither \\
Nested Bridge & 4 & Comparison of the end values of two bridge paths \\
\bottomrule
\end{tabularx}
\end{table}

\subsection{Construction and Validation}
\label{subsec:validation}
\textbf{Instance generation.}
For each template, we sample the required facts \(F\) from \(\mathcal{KG}\), generate the question \(q\) and valid answers \(A\), and then assign each fact \(f_i\) to its Wikipedia document \(d_i=\Gamma(f_i)\).

\textbf{Validation and repair.}
We verify structural constraints rule-based and use \textsc{GPT-OSS-120B} for semantic assessments and leakage checks, to satisfy the requirements defined in \Cref{subsec:requirementAnalysis}. Invalid instances are repaired without modifying the source documents and then fully validated again. Candidates that still fail are discarded. Validation details and repair operations are provided in \Cref{app:validation}.

\textbf{Distractors.}
We retrieve topically related Wikipedia documents with \textsc{sentence-transformers/all-MiniLM-L6-v2} and retain a candidate \(d\) only if it supports none of the required facts:
\[
\operatorname{supp}(d,f)=0
\qquad \forall f\in F.
\]

\subsection{\texorpdfstring{\protect\fondueicon~\fondue~Benchmark\hspace{0.3em}}{\fondue Benchmark}}
\label{sec:fondueqa}
 
Our final release comprises \textbf{13{,}063 questions} with 20 context documents each: 39.3\% Bridge, 39.3\% Comparison, 9.8\% Contrastive, 9.8\% Cross-Property, and 1.7\% Nested Bridge. Answers are yes/no (36.5\%), option selection (21.8\%), or open (41.6\%). Each required fact is annotated with the Wikipedia pageviews of its role entity (median 9{,}778, interquartile range 2{,}407--55{,}261), which we use in \Cref{sec:eval-drivers} as a proxy for parametric accessibility. Further details on the popularity measure and Wikipedia corpus are given in \Cref{app:pop_wiki}. \Cref{tab:fondueqa-core-stats} and \Cref{app:dataset-statistics} give the composition and the filtering breakdown.
\fondue ~is publicly available as a versioned benchmark at \url{https://anonymous.4open.science/r/FondueQA-7C6B}, with quarterly releases to provide new benchmark instances over time.

\section{Experimental Setup}
\label{sec:experimental_setup}
\textbf{Language Models and Configurations.}
We evaluate models from four families spanning different parameter scales, dense and mixture-of-experts architectures, and where supported, both standard and reasoning inference. For models supporting both modes, we use the model's native mechanism for activating reasoning. \Cref{tab:model_catalogue} in \Cref{app:model_details} summarizes the evaluated configurations.

All models receive the same prompt, which explicitly permits to draw on parametric knowledge and instructs to return a concise final answer. We set the temperature to \(0\) and top-\(p\) to \(0.9\), with a maximum reasoning budget of 15,000 tokens where applicable and a 150-token final-answer budget otherwise. Standard and reasoning runs differ only in the activation of reasoning mode. We report complete prompts, inference parameters, and serving details in \Cref{app:model_details}.

\textbf{Datasets and Context Conditions.}
We use extended versions of established benchmarks, denoted by the \ext\ suffix, which are subject to the same checks as \fondue\ described in \Cref{subsec:validation}. We evaluate on \fondue, \twowikiext, \hotpotext, and \mintqaext. Our primary analyses use \fondue, whose construction and re-balancing is explicitly designed to fulfill the requirements introduced in \Cref{subsec:requirementAnalysis}. Applying these constraints post-hoc to the \ext\ variants can alter the composition of the retained subsets. We therefore use \ext\ datasets primarily to test whether the observed effects generalize beyond \fondue. Conversion of the \ext\ variants is reported in \Cref{app:external-datasets}, with results in \Cref{app:external-dataset-results}.

Following our definition of reasoning paths (\Cref{def:reasoningPath}) and evidence gaps (\Cref{def:evidenceGap}), we evaluate a complete reasoning path, a complete gap with all gold evidence removed, and targeted partial gaps obtained by removing individual facts. A separate question-only condition removes the retrieval context entirely. For each removed fact, we additionally query the atomic fact in isolation to test whether the fact is available from the model's parametric knowledge. At a fixed evidence gap, we further evaluate reasoning traps (\Cref{def:reasoningTrap}) by varying the number of type-compatible distractors that provide plausible alternative reasoning paths. Construction details and coverage are provided in \Cref{app:reasoning-traps}. In addition to evidence availability, we control for positional effects by keeping retained gold-document positions fixed for direct role comparisons. The resulting context conditions and positional controls are detailed in \Cref{app:context-conditions}.

\textbf{Evaluation Metrics.}
We use \textit{Exact Match} (EM) as metric across all question formats and additionally define token-level F1 for open-ended answers. For a model prediction \(p\) and the set of valid answer forms \(A\) defined in \Cref{subsec:formalSetting}, including the canonical answer and permissible aliases, we compute
\[
\mathrm{EM}(p,A)=\max_{a\in A}~\mathrm{EM}(p,a), \qquad
\mathrm{F1}(p,A)=\max_{a\in A}~\mathrm{F1}(p,a).
\]
We score the single-hop fact questions using the same answer normalization and alias matching as the corresponding multi-hop questions, as detailed in \Cref{app:evaluation-details}.
\section{Findings}
\label{sec:findings}
\textbf{Bridge Evidence Can Outweigh Answer-Bearing Evidence.}\label{sec:eval-drivers} On \fondue, retaining the bridge fact often outweighs retaining the answer-bearing target document, particularly for smaller models. This shows that under partial retrieval, parametric knowledge can compensate for the missing fact, but its effectiveness may depend on the structural role left out.

\begin{figure}[t]
    \centering
    \includegraphics[height=0.25\textheight,keepaspectratio]{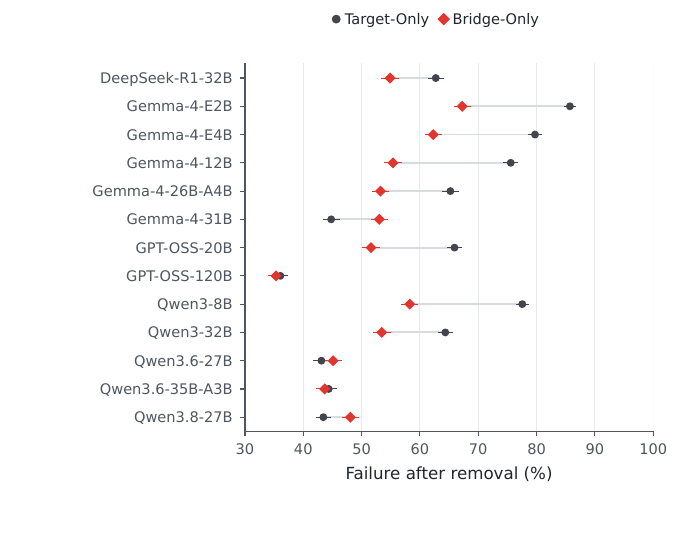}
    \hspace{0.015\textwidth}
    \includegraphics[height=0.25\textheight,keepaspectratio]{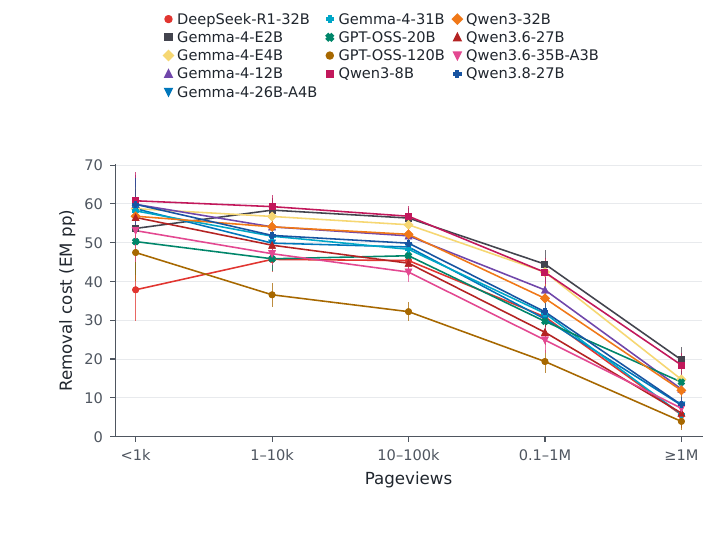}
    \caption{
    \textbf{Evidence structure and popularity shape compensation.} Left: Failure rates under role-specific evidence gaps for position-matched bridge questions. 
    Right: Performance loss $\Delta_{\mathrm{removal}}(f_i) = \mathrm{EM}_{\mathrm{full}}-\mathrm{EM}_{-f_i}$ as a function of entity popularity. Error bars show 95\% bootstrap confidence intervals.
    }
    \label{fig:drivers}
\end{figure}

For bridge questions, we compare two evidence gaps for a reasoning path $\pi=(f_1, f_2)$ within the same question. In \tmpbridgeonly,
the first hop $f_1=(e_0,r_1,e_1)$ is retained, thereby identifying the intermediate entity \(e_1\). In  \tmptargetonly, 
the second hop $f_2=(e_1,r_2,e_2)$ is retained, which directly contains the final answer $a{=}e_2$. Although \tmptargetonly directly contains the answer-bearing fact, \tmpbridgeonly often performs better, particularly for smaller models (\Cref{fig:drivers}). The predominant asymmetry persists under position-matched conditions, as detailed in \Cref{app:position-matching}. 
We argue that bridge evidence is particularly valuable, as it explicitly identifies the relevant intermediate entity and thereby determines to which entity missing knowledge must be applied. Although an answer-bearing target document contains the final fact, it could be less immediately helpful if it first requires reconstructing which of the documents actually pertains to the original question.

\textbf{Popularity in Multi-Hop QA.} Across evidence gaps, models and question types, missing facts involving more popular entities incur lower removal costs, suggesting more successful parametric compensation.

We use Wikipedia page-views as a proxy for parametric accessibility and define the removal cost of a fact \(f\) as
$
\Delta_{\mathrm{removal}}(f)
=
\mathrm{EM}_{\mathrm{full}}
-
\mathrm{EM}_{-f}.
$
Details on the popularity measure are provided in \Cref{app:pop_wiki}. As shown in the right panel of \Cref{fig:drivers} missing evidence becomes less costly for more popular entities, indicating easier parametric compensation. Models can thus more frequently compensate for missing facts about popular entities, while external evidence remains more important for rarer entities. This extends the popularity effect observed for single-hop QA by \citet{mallen-etal-2023-trust} to multi-hop questions under partial retrieval.

\textbf{Atomic Recall Does Not Guarantee Compensation.}\label{sec:eval-limits} 
Atomic recall and compensation are related but not equivalent. Models can answer correctly without recalling the missing fact in isolation, while also failing to answer correctly despite recalling it in isolation.

\begin{figure}[t]
    \centering
    \includegraphics[height=0.25\textheight,keepaspectratio]{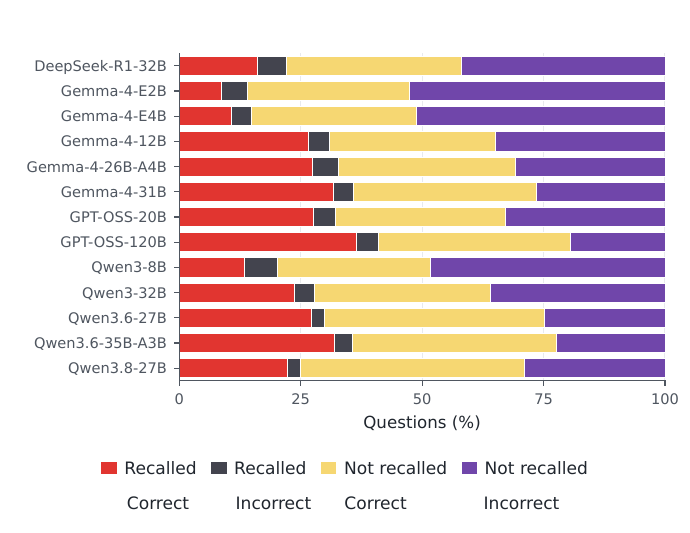}
    \hspace{0.015\textwidth}
    \includegraphics[height=0.25\textheight,keepaspectratio]{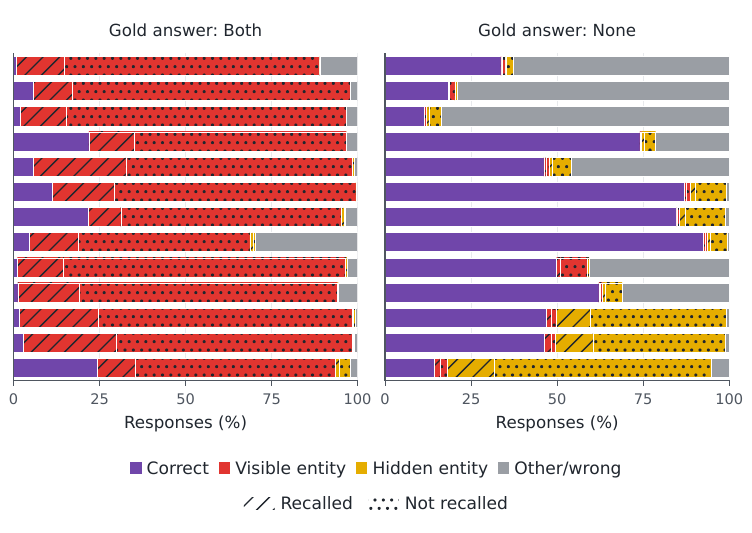}
    \caption{
    \textbf{Parametric availability and contextual behavior under partial evidence.} Left: Joint outcomes of atomic recall of the missing fact and correctness of the corresponding multi-hop question under partial evidence. Right: Contrastive questions with partial evidence systematically shift response behavior toward the evidence-supported entity.
    }
    \label{fig:partial-evidence-behavior}
\end{figure}

The left panel of \Cref{fig:partial-evidence-behavior} relates this atomic recall to the correctness of the corresponding multi-hop question under partial evidence. Cases in which the missing fact is recalled but the multi-hop answer remains incorrect occur only to a limited extent in most models, but are present across the board. Conversely, many models answer a significant proportion of the questions correctly even though the missing fact is not recalled atomically. Across several models, this proportion is approximately \(32\)–\(45\%\). Therefore, a single failed recall does not preclude successful compensation. Since, by design, the missing fact is not contained in the remaining context, the context cannot provide it directly. However, it could, for example, highlight related entities and/or narrow down the plausible solution space, thereby guiding the model towards the correct answer.

Contrastive questions allow for a more detailed examination of this behavior. They are formulated as a choice between two explicitly named entities, although, depending on the underlying facts, \emph{Both} or \emph{None} may also be the correct answer. Under partial evidence, we refer to the entity whose relevant evidence is still supported in the context as a \emph{visible entity}; the entity whose evidence has been removed as a \emph{hidden entity}. The right panel of \Cref{fig:partial-evidence-behavior} shows the response distribution for gold answers \emph{Both} and \emph{None} and additionally distinguishes whether the fact of the entities was recalled atomically.

When Gold=\emph{Both}, the models often respond only with the visible entity. This is particularly plausible when the hidden fact is not recalled. It is important to note, however, that visible-entity responses also occur even when atomic recall is successful and account for up to about \(27\%\) of the responses in some models. In these cases, the missing information can be recalled in isolation but is not combined with the visible evidence to produce the \emph{Both} response. In contrast, when Gold=\emph{None}, visible-entity responses are rare. When models select a single entity, the response more frequently falls on the hidden entity and accounts for about half of the predictions in some models. Both recalled and not-recalled cases occur here, so atomic recall only partially explains the response behavior in this scenario as well.

\textbf{Competing Evidence Can Shift Parametric Compensation.}\label{sec:candidate_competition} Under a fixed evidence gap, plausible alternative entities can make it harder to use the given answer-bearing evidence. Moreover, this effect depends on what the competing evidence supports, with answer-consistent candidates strengthening the correct answer and consistently misleading documents shifting models away from it.

\begin{figure}[t]
    \centering
    \includegraphics[height=0.25\textheight,keepaspectratio]{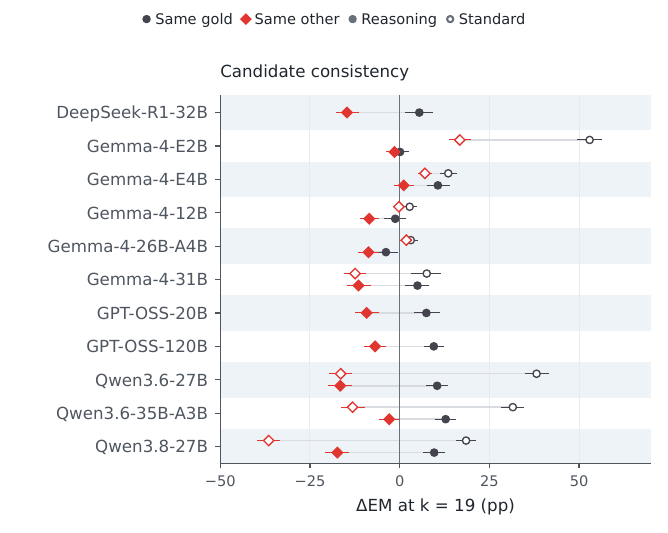}
    \hspace{0.015\textwidth}
    \includegraphics[height=0.25\textheight,keepaspectratio]{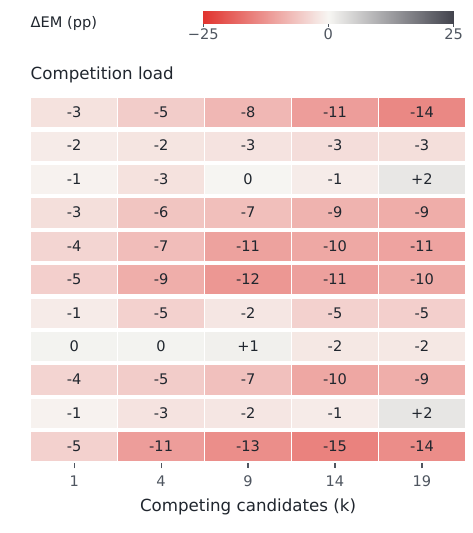}
    \caption{\textbf{Candidate competition.} Left: $\Delta$EM at $k=19$ across candidate-consistency conditions. Right: $\Delta$EM with increasing numbers of random candidates in reasoning-mode.
    }
    \label{fig:candidate-competition}
\end{figure}

\begin{figure}[t]
    \centering
    \includegraphics[height=0.02\textheight]{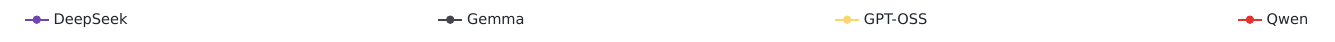}
    \includegraphics[height=0.23\textheight,keepaspectratio]{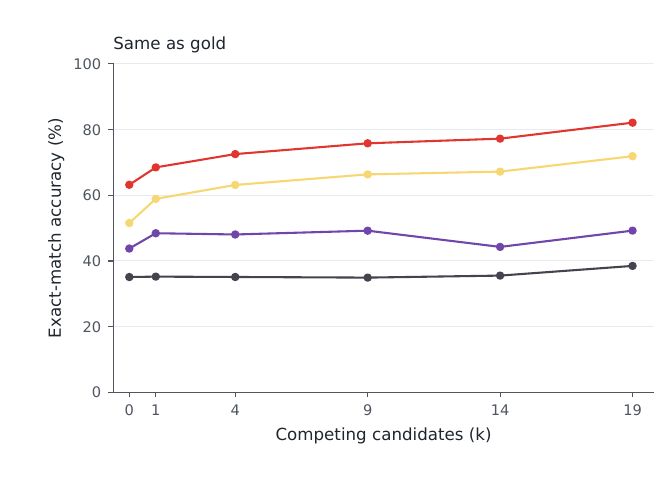}
    \hspace{0.015\textwidth}
    \includegraphics[height=0.23\textheight,keepaspectratio]{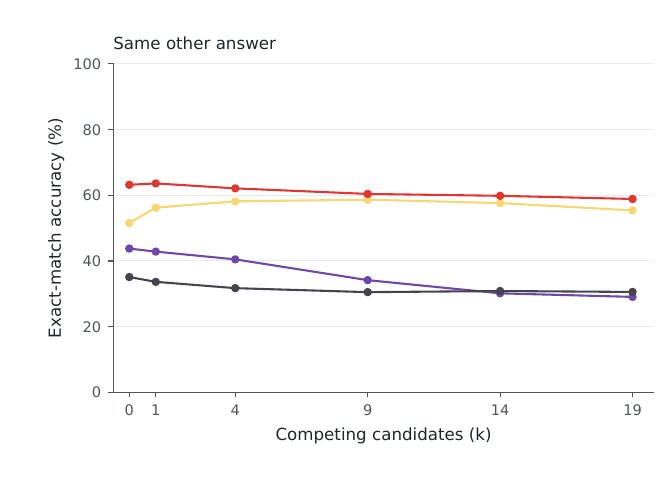}
    \caption{
    \textbf{Competing evidence.}
    Exact Match with increasing number of candidate documents. Left: \emph{Same-Gold} candidates vs. Right: \emph{Same-Other} candidates.}
    \label{fig:candidate-steering}
\end{figure}

The right-hand plot in Figure~\ref{fig:candidate-competition} shows that, in reasoning-mode, additional plausible candidates make it more difficult for several models to utilize the answer-bearing evidence. The effect is heterogeneous across models, but can become substantial as competition increases, with drops of up to about 15 percentage points for some configurations.
Candidate competition thus represents a specific reasoning trap: multiple semantically plausible bindings compete with the path intended for the question. To determine whether this effect is solely attributable to additional distractors or generally higher context complexity, we control the semantic direction of the candidate evidence. In a \emph{Same-Gold} setting, the candidate documents support the same answer as the correct target path (e.g. other directors from 'Canada'), thereby generating additional answer-consistent paths. In the \emph{Same-Other} setting, all documents consistently support another plausible but incorrect alternative (e.g. directors with a different nationality). The direction of this effect changes with candidate consistency. At $k{=}19$, several configurations show shifts of more than $+10$ percentage points, while non-reasoning modes can even reach roughly $+50$ points for \emph{Same-Gold} and $-35$ points for \emph{Same-Other}. The family-level trajectories in \Cref{fig:candidate-steering} show the same tendency as competition increases for reasoning-mode configurations, although the magnitude varies across model families. Additional model-level results are provided in \Cref{app:candidate-competition-results}.

\textbf{Implications for Retrieval-Augmented QA.}
Our findings suggest that partial retrieval is not uniformly harmful. Retrieval should not only account for coverage, but also for how easily missing information can be compensated for and which role the retrieved evidence plays in the reasoning path. In particular, less popular facts may require stronger retrieval support, while bridge evidence can sometimes be more useful than answer-bearing evidence. At the same time, retrieving many plausible alternative entities can be harmful when the bridge is missing, suggesting that candidate ambiguity matters alongside recall. Finally, models should be able to reliably integrate parametrically available facts when evidence is missing anyway.
\section{Conclusion}
We introduce \fondue, a multi-hop QA benchmark in which every required fact is uniquely localized to a single gold document and verified to be absent from the remaining context. This enables controlled document-level evidence ablations to study parametric compensation under incomplete retrieval. Our results show that successful compensation depends not only on whether information is available parametrically, but also on the structure and ambiguity of the remaining evidence. \fondue ~thereby provides a controlled setting for studying how parametric knowledge interacts with incomplete retrieval in multi-hop reasoning.

\clearpage
\subsection*{AI use statement}

In this work generative AI tools were used to provide feedback on research methodology, assist with method implementation and code development, create and edit research artifacts and tables, summarize existing literature, identify potential research gaps and relevant literature, support brainstorming, improve the readability of the manuscript and suggest keywords.

All AI-assisted outputs were reviewed and verified by the authors. AI-assisted code and research artifacts were checked for correctness, literature-related suggestions were verified against the original sources, and all manuscript edits were reviewed before inclusion. The authors take full responsibility for the final content, claims, results and artifacts presented in this work.

\subsection*{Ethics statement}

\benchmark{} is constructed from publicly available information in Wikipedia and Wikidata and therefore includes questions and facts concerning real persons. As both resources are collaboratively maintained and continuously editable, their contents may contain inaccuracies, outdated information, biases, or, in rare cases, deliberately misleading, discriminatory, offensive, or otherwise harmful statements. Such content is neither intended nor endorsed by the authors and should not be interpreted as independently verified claims. During benchmark construction, we apply validation and filtering procedures to remove inconsistent, unsupported, ambiguous, or otherwise unsuitable instances, including content that appears inappropriate for inclusion. However, these procedures cannot guarantee the factual correctness or harmlessness of every source statement. Consequently, \benchmark{} should be treated as a research benchmark derived from these sources rather than as an authoritative source of biographical or factual information about individuals, and users should consider the provenance and limitations of the underlying data when interpreting examples concerning real persons.

\subsection*{Reproducibility statement}

We provide the benchmark, validated external benchmark variants, construction and evaluation code, validation prompts and experiment configurations in the anonymous repository at \url{https://anonymous.4open.science/r/FondueQA-7C6B}. The appendix documents the benchmark construction and validation procedure (\Cref{app:validation,app:dataset-statistics}), models and serving configuration (\Cref{app:model_details}), context conditions (\Cref{app:context-conditions}) and external benchmark adaptation (\Cref{app:external-datasets}). All experiments used a fixed seed 42 and the released configurations reproduce the experimental conditions reported in the paper.

\subsubsection*{Acknowledgments}
We acknowledge the funding support from the \emph{Bayerisches Landesamt für Steuern} for the Bavarian AI Taxation Laboratory. Moreover, we gratefully acknowledge the scientific support and HPC resources provided by the Erlangen National High Performance Computing Center (NHR@FAU) of the Friedrich-Alexander-Universität Erlangen-Nürnberg (FAU) under the NHR project DeepSmall - Meta-learning for regularizing deep networks under small data regimes. NHR funding is provided by federal and Bavarian state authorities. NHR@FAU hardware is partially funded by the German Research Foundation (DFG) – 440719683.
\clearpage

\bibliography{iclr2027_conference}
\bibliographystyle{iclr2027_conference}

\appendix

\section{Benchmark Construction Details}
\label{app:benchmark_details}

\subsection{Question Templates}
\label{app:templates}

\Cref{tab:question_templates} summarizes the question families used in \fondue. The examples illustrate the corresponding reasoning structures but are not drawn from the original dataset.

\begin{table*}[htbp]
    \caption{\textbf{Question families in \benchmark{}.}
    Representative examples illustrating the reasoning structure and typical answer form of each question type.}
    \centering
    \setlength{\tabcolsep}{5pt}
    \label{tab:question_templates}
    \begin{tabularx}{\linewidth}{
        L{0.14\linewidth}
        X
        L{0.18\linewidth}
    }
        \toprule
        \textbf{Type} &
        \textbf{Example question} &
        \textbf{Example answer} \\
        \midrule

        Bridge &
        In which country is the director of the movie \emph{Titanic} born? &
        Canada \\

        \addlinespace

        Comparison &
        Do Barack Obama and George W. Bush have the same occupation? &
        Yes \\

        \addlinespace

        Contrastive &
        Who was born in Honolulu, Barack Obama or George W. Bush? &
        Barack Obama \\

        \addlinespace

        Cross-property &
        Is the director of \emph{Titanic} also the screenwriter of
        \emph{Avatar}? &
        Yes \\

        \addlinespace

        Nested Bridge &
        Who is older, the director of \emph{Titanic} or the director of
        \emph{Jaws}? &
        Steven Spielberg; Director of \emph{Jaws} \\

        \bottomrule
    \end{tabularx}
\end{table*}

\subsection{Validation and Repair}
\label{app:validation}

Our validation pipeline operationalises the requirements for controlled evidence removal introduced in \Cref{subsec:requirementAnalysis}. An instance is retained only if every required fact can be uniquely assigned to its designated gold document and all applicable question-, evidence-, and context-level checks are satisfied.

\paragraph{Gold support and fact isolation.} For every required fact \(f_i\), we verify that its assigned gold document \(\Gamma(f_i)\) supports the fact and that no other context document provides the same required evidence. This directly enforces conditions (C1) and (C2).

\paragraph{Question and entity validation.} We verify that the question does not directly expose a required fact or reveal which answer candidate is correct, corresponding to condition (C3). We further check that the entities required to follow the intended reasoning structure are identified unambiguously. Questions that remain ambiguous or admit an unintended answer path are rejected or, where applicable, repaired.

\paragraph{Answer and structural validation.} We verify that the designated evidence supports the composed answer and that the instantiated reasoning structure matches the intended template. In particular, required hops must contribute distinct evidence; instances with redundant hops or unintended single-document solution paths are rejected or repaired.

\paragraph{Distractor and context validation.} Candidate distractors are retained only if they contain none of the required facts. After assembling the complete context, we perform a final document-level leakage check and reject contexts containing residual required evidence, duplicate documents, or other violations of evidence isolation. We additionally enforce the context-length constraints used during benchmark construction.

\paragraph{Automatic adjudication.} We combine deterministic checks with semantic adjudication by GPT-OSS-120B. Deterministic checks are used whenever the relevant constraint can be established directly from the benchmark structure or annotations, while semantic checks are used for support and leakage judgments that require natural-language interpretation. The validation procedure is intentionally conservative: ambiguous cases are rejected rather than retained when evidence is insufficient to establish the required isolation conditions. The complete validation prompts and implementation are provided in the anonymous repository at \url{https://anonymous.4open.science/r/FondueQA-7C6B}.

\paragraph{Repair operations.}
\label{app:repair}
If a bridge-question instance $x=(q,D,A,F,\Gamma)$ violates the fact isolation, we attempt a conservative repair
\[
\mathcal{T}:x\mapsto x'=(q',D,A',F',\Gamma'),
\]
where the original Wikipedia documents $D$ remain unchanged. For a reasoning path $\pi=(f_1,\ldots,f_k)$, we consider four transformations:
\[
\begin{aligned}
\mathcal{T}_{\mathrm{retarget}} &:~
(f_1,\ldots,f_k)
\mapsto
(f_1,\ldots,f_{k-1},f_k'),\\
\mathcal{T}_{\mathrm{flip}} &:~
(f_1,\ldots,f_k;\Gamma)
\mapsto
(f_k',\ldots,f_1';\Gamma'),\\
\mathcal{T}_{\mathrm{compress}} &:~
(f_1,f_2,f_3)
\mapsto
(f_1',f_2'),\\
\mathcal{T}_{\mathrm{reframe}} &:~
q(e_0)
\mapsto
q(\delta(e_0)).
\end{aligned}
\]
Here, \emph{retargeting} replaces the terminal fact with one supported exclusively by its designated gold document; \emph{flipping} constructs a valid reasoning path with reversed gold-document roles and updates $\Gamma$ accordingly; \emph{compression} replaces a path containing a redundant hop with a shorter valid path; and \emph{reframing} replaces an explicitly named source entity $e_0$ in the question with a unique, non-answer-bearing description $\delta(e_0)$. Any resulting changes to the required facts or answer are reflected in $F'$ and $A'$. A repaired instance is retained only if it satisfies all validation conditions again.

\paragraph{Final acceptance.}
Repairs never modify the underlying source documents. After repair, the full instance, including the assembled context and distractors, is passed through the complete validation procedure again. Instances that still violate any required condition are discarded.

\subsection{Dataset Statistics and Filtering}
\label{app:dataset-statistics}

\begin{figure}
    \centering
    \includegraphics[width=0.95\linewidth]{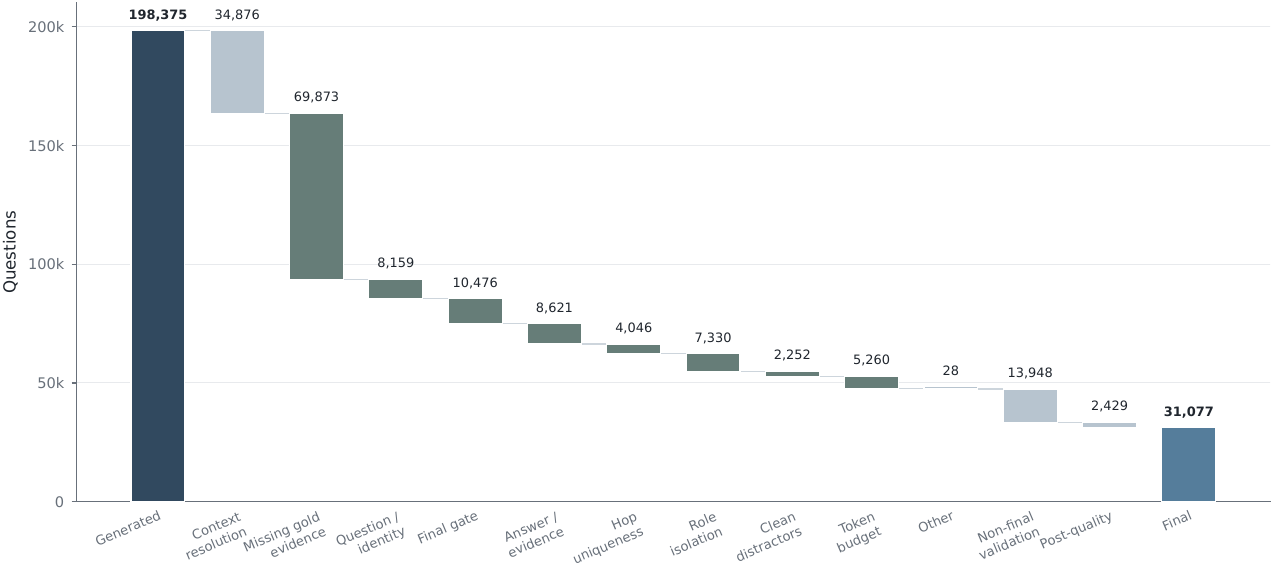}
    \caption{\textbf{\fondue\ construction filtering.} Intermediate bars show the number of candidate questions removed by each validation or filtering step, while the first and final bars report the initial candidate pool and resulting validated dataset size.}
    \label{fig:construction-waterfall}
\end{figure}

\textbf{Balancing.} Starting from 198,375 generated candidates, 31{,}077 instances pass validation and repair  (\Cref{fig:construction-waterfall}). Because chained structures must satisfy more conditions jointly, the validated pool is skewed toward Comparison, with substantially fewer Bridge instances surviving validation. We therefore rebalance the four main question types by retaining all validated Bridge questions and stratified undersampling the remaining types to a 40/40/10/10 target distribution for Bridge, Comparison, Contrastive, and Cross-property questions, respectively. Finally, we add all validated Nested Bridge instances, which are too sparse for meaningful undersampling, yielding the 13{,}063-question paper release summarized in \Cref{tab:fondueqa-core-stats}.

\begin{table}[htbp]
    \caption{\textbf{\fondue\ construction pool and rebalanced paper release.}
    Percentages and distributional statistics refer to the final 13,063-question release.}
    \centering
    \small
    \setlength{\tabcolsep}{4pt}
    \label{tab:fondueqa-core-stats}
    \begin{tabular}{lr}
        \toprule
        Statistic & Value \\
        \midrule
        Generated candidates & 198,375 \\
        Candidates with complete gold context & 163,499 \\
        Validated construction pool & 31,077 \\
        Published paper release & 13,063 \\
        Validated release backlog & 18,014 \\
        Documents per question & 20 \\
        \midrule
        Bridge & 5,138 (39.3\%) \\
        Comparison & 5,138 (39.3\%) \\
        Contrastive & 1,285 (9.8\%) \\
        Cross-property & 1,284 (9.8\%) \\
        Nested bridge & 218 (1.7\%) \\
        \midrule
        Yes/no answers & 4,771 (36.5\%) \\
        Option-selection answers & 2,854 (21.8\%) \\
        Open answers & 5,438 (41.6\%) \\
        \midrule
        Context tokens, median [IQR] & 9,142 [5,966--13,352] \\
        Maximum context tokens & 29,734 \\
        Role-entity pageviews, median [IQR] & 9,778 [2,407--55,261] \\
        \bottomrule
    \end{tabular}
\end{table}

\subsection{Wikipedia Corpus and Entity Popularity}
\label{app:pop_wiki}

\paragraph{Wikipedia corpus.} We use an English Wikipedia snapshot from October 2017 as the document corpus. The snapshot intentionally predates the knowledge cutoffs of modern language models considered in our experiments by a substantial margin. Since our analysis studies parametric compensation, this provides at least the possibility that the entities and facts contained in the benchmark could have occurred in pretraining data. Benchmark facts are sampled from current Wikidata, but an instance is retained only if the corresponding fact is supported by its assigned document in the 2017 Wikipedia corpus. Facts that are too recent, have changed over time, or are otherwise unsupported by the historical document therefore fail validation and are discarded. Such temporal mismatches may contribute to the \emph{Missing gold evidence} removals reported in \Cref{fig:construction-waterfall}.

\paragraph{Entity popularity.} Following \citet{mallen-etal-2023-trust}, we use Wikipedia pageviews as a proxy for entity popularity and, indirectly, for the frequency with which an entity may have appeared in pretraining data. For each entity, we sum its monthly Wikipedia pageviews over all months of 2023. We choose a period close to, but preceding the relevant training cutoffs of the evaluated models where such information is available. Compared with more recent pageview statistics, this potentially reduces the risk of assigning high popularity to entities that became prominent only after the model's training period. Pageviews remain only a proxy for parametric accessibility and should not be interpreted as direct evidence of pretraining exposure.

\section{Experimental Details}
\label{app:experimental_details}

\subsection{Model Serving}
\label{app:model_details}
\begin{table*}[htbp]
\caption{\textbf{Language models evaluated in our experiments.}
We evaluate standard and reasoning inference whenever supported by the model and our serving configuration. For mixture-of-experts models, parameter counts are reported as total / active parameters.}
\label{tab:model_catalogue}
\begin{center}
\fondueTableSetup
\begin{tabularx}{\linewidth}{
    L{0.11\linewidth}
    L{0.29\linewidth}
    L{0.18\linewidth}
    L{0.13\linewidth}
    C{0.10\linewidth}
    C{0.10\linewidth}
}
\toprule
\textbf{Family} & \textbf{Model} & \textbf{Parameters} &
\textbf{Architecture} & \textbf{Standard} & \textbf{Reasoning} \\
\midrule

\multirow[c]{4}{*}{\textit{Qwen}}
& Qwen3.8-27B & 27B & Dense & \checkmark & \checkmark \\
& Qwen3.6-35B-A3B & 35B / 3B active & MoE & \checkmark & \checkmark \\
& Qwen3.6-27B & 27B & Dense & \checkmark & \checkmark \\
& Qwen3-8B & 8B & Dense & \checkmark & \checkmark \\
& Qwen3-32B & 32B & Dense & \checkmark & \checkmark \\
\midrule

\multirow[c]{5}{*}{\textit{Google}}
& Gemma-4-E2B-IT & 2B & Dense & \checkmark & \checkmark \\
& Gemma-4-E4B-IT & 4B & Dense & \checkmark & \checkmark \\
& Gemma-4-12B-IT & 12B & Dense & \checkmark & \checkmark \\
& Gemma-4-26B-A4B-IT & 26B / 4B active & MoE & \checkmark & \checkmark \\
& Gemma-4-31B-IT & 31B & Dense & \checkmark & \checkmark \\
\midrule

\multirow[c]{2}{*}{\textit{OpenAI}}
& gpt-oss-20b & 20B / 3.6B active & MoE & \xmark & \checkmark \\
& gpt-oss-120b & 120B / 5.6B active & MoE & \xmark & \checkmark \\
\midrule

\textit{DeepSeek}
& DeepSeek-R1-Distill-Qwen-32B & 32B & Dense & \xmark & \checkmark \\

\bottomrule
\end{tabularx}
\end{center}

\end{table*}

\paragraph{General prompt.} All models receive the same evaluation system prompt across context conditions.

\begin{quote}
\small
\ttfamily
\raggedright

Answer the question based on the provided context paragraphs if possible. If an information is not present in the context you may use your internal parametric knowledge. You must perform multi-hop reasoning by combining information from multiple documents if necessary.

\medskip

Instructions:

1. Identify the supporting facts from the context (if possible) that are required to answer the question.

2. Synthesize the information to formulate the final answer.

3. Keep your answer concise and factual.

\medskip

Give a short, direct answer, usually one word or a short phrase.

For yes/no questions, answer only yes or no.

Do not explain your reasoning.

\end{quote}

\paragraph{Question only prompt.}
When no context is provided, the model receives only the question together with the following system prompt.

\begin{quote}
    \small
    \ttfamily
    \raggedright

You are a precise question-answering assistant.
Answer the question.
Give a short, direct answer, usually one word or a short phrase.
For yes/no questions, answer only yes or no.
Do not explain your reasoning.
\end{quote}

\paragraph{Inference parameters.}
We use temperature \(0\) with \(p=0.9\) for all model configurations. Standard inference is limited to 150 output tokens. For models with a native reasoning mode we allow a maximum reasoning budget of 15,000 tokens. All runs use a fixed random seed of 42. We use a maximum context budget of 40,000 tokens.

\paragraph{Model serving.}
We serve all models locally using vLLM (version: 0.26.0) on NVIDIA H100 GPUs. Runs use up to four GPUs with tensor parallelism of size 4, and pipeline parallelism of size 1. Model-specific reasoning modes are activated through their native inference mechanisms where supported. The models used are reported in
\Cref{tab:model_catalogue}.

\subsection{Extended Benchmark Construction}
\label{app:external-datasets}

\paragraph{External benchmark adaptation.}
For \hotpot\ and \twowiki, we retain the original question, answer, and gold documents and apply our pipeline post-hoc to the provided contexts. \mintqa\ requires an additional context-construction step because the source instances do not provide a document context. We therefore resolve the annotated gold entities to their corresponding documents and construct the context before applying the same validation procedure. Instances for which a required gold document cannot be resolved are discarded. Since our analysis targets multi-hop reasoning, we additionally exclude \mintqa\ instances that contain only a single reasoning hop.

\paragraph{Validation procedure.}
For each remaining source instance, we resolve the annotated supporting evidence and map the required facts to their corresponding gold documents and reasoning roles. We then apply the same checks used for \fondue: each required fact must be supported by its assigned gold document, must not be recoverable from another context document, and must not be directly exposed by the question. We additionally verify that the composed answer is supported, screen distractor documents for leakage of required facts, and enforce the applicable context-length constraints.

\paragraph{No QA repair.}
In contrast to the construction of \fondue, we treat the original question-answer instances of the external benchmarks as immutable. We do not rewrite questions, answers, reasoning chains, or gold-document text to make an instance satisfy the requirements. If a content-level violation cannot be resolved without changing the original QA instance, the instance is discarded. The only modification permitted is the replacement of a distractor document that leaks required evidence with a clean distractor.

\paragraph{Conservative filtering.}
Our validation procedure is intentionally conservative. A rejection indicates that an instance did not satisfy a required check with sufficient confidence under our pipeline; it should not be interpreted as a definitive claim that the original benchmark instance is incorrect. Validation checks are applied sequentially, and an instance is removed at the first failing stage. Consequently, the reported rejection category denotes the first observed failure and does not imply that later checks would necessarily have been satisfied. In ambiguous cases, we prefer to discard an instance rather than retain one that may violate a check. The reported retention rates therefore characterize the subset for which our evidence-control requirements can be established reliably, rather than the overall quality of the source benchmarks. \Cref{tab:external_drop_reasons} presents the filtering outcomes for all benchmarks.

Applying the validation to \musique\ leaves only a small number of valid instances. We therefore exclude \musiqueext\ from the main cross-benchmark analysis and focus on \mintqaext, \hotpotext, and \twowikiext.

\begin{table*}[htbp]
    \caption{
    \textbf{Filtering outcomes for the extended benchmarks.}
    \mintqa\ additionally requires exclusion of
    single-hop instances and document resolution before entering the common
    validation pipeline. Retention rates in parentheses are computed relative
    to the initial candidate pool.
    }
    \centering
    \label{tab:external_drop_reasons}

    \begin{tabularx}{\textwidth}{Xrrrr}
        \toprule
        \textbf{Outcome}
        & \makecell{\textbf{HotpotQA}\\\textbf{-Ext}}
        & \makecell{\textbf{2WikiMultiHopQA}\\\textbf{-Ext}}
        & \makecell{\textbf{MINTQA}\\\textbf{-Ext}}
        & \makecell{\textbf{MuSiQue}\\\textbf{-Ext}} \\
        \midrule

        Initial candidates
        & 7,404
        & 12,576
        & 17,887
        & 2,417 \\

        \midrule
        Pre-validation: single-hop
        & --
        & --
        & 5,894
        & -- \\

        Pre-validation: \\document resolution
        & --
        & --
        & 2,592
        & -- \\

        Entered validation
        & 7,404
        & 12,576
        & 9,401
        & 2,417 \\

        \midrule
        Evidence mapping
        & 414
        & 224
        & 0
        & 192 \\

        Missing Evidence
        & 130
        & 1,756
        & 1,652
        & 470 \\

        Question / identity
        & 2,353
        & 27
        & 20
        & 338 \\

        Answer / evidence
        & 341
        & 88
        & 10
        & 62 \\

        Role isolation
        & 680
        & 2,677
        & 3,822
        & 566 \\

        Clean distractors
        & 11
        & 2
        & 80
        & 1 \\

        Token budget
        & 0
        & 0
        & 140
        & 0 \\

        \midrule
        Total rejected
        & 3,929
        & 4,774
        & 14,210
        & 1,629 \\

        Pending / technical
        & 51
        & 62
        & 490
        & 257 \\

        \midrule
        \textbf{Retained}
        & \textbf{3,424 (46.2\%)}
        & \textbf{7,740 (61.5\%)}
        & \textbf{3,187 (17.8\%)}
        & \textbf{531 (22.0\%)} \\

        \bottomrule
    \end{tabularx}
\end{table*}

\subsection{Context Conditions}
\label{app:context-conditions}

Our evaluation contexts factorize two independent dimensions: which gold evidence is available and where the retained gold documents are placed. \Cref{tab:context_conditions} summarizes the resulting conditions.

\paragraph{Evidence availability.}
We evaluate the complete gold context, contexts with no gold evidence, and partial-evidence conditions in which only selected reasoning roles are retained or individual gold documents are removed. A separate question-only condition removes the retrieval context entirely. These conditions induce the Evidence Gaps used throughout our analyses.

\paragraph{Evidence placement.}
Contexts in \fondue\ contain 20 documents. Depending on the condition, retained gold documents are placed toward the front, middle, or end of the context. The exact positions are given in \Cref{tab:gold_positions}.

\paragraph{Position-matched comparisons.}
For direct comparisons between evidence roles, we keep the positions of the retained gold documents fixed across conditions. This ensures that differences between, for example, Bridge-only and Target-only contexts are not driven by systematic differences in document position.

\begin{table*}[htbp]
\caption{\textbf{Factorized context construction.}
Each evaluation context combines an evidence-availability setting with a compatible evidence-placement setting. Availability determines which gold documents are present; placement determines their positions in the 20-document context.}
\label{tab:context_conditions}
\begin{center}
\fondueTableSetup
\begin{tabularx}{\linewidth}{
    L{0.16\linewidth} Y
    @{\hspace{0.8cm}}
    L{0.16\linewidth} Y
}
\toprule
\multicolumn{2}{c}{\textbf{Evidence availability}}
&
\multicolumn{2}{c}{\textbf{Evidence placement}}
\\
\cmidrule(r){1-2}\cmidrule(l){3-4}
\textbf{Setting} & \textbf{Gold evidence present}
&
\textbf{Setting} & \textbf{Placement}
\\
\midrule

Question only & None
&
Front & Starts at position 1
\\

No gold & None; distractors only
&
Middle & Centered at position 10
\\

All gold & All gold documents
&
End & Ends at position 19
\\

Role only & One selected role
&
&
\\

Leave-one-out & All except one
&
&
\\

\midrule
\multicolumn{4}{c}{
\(\text{Evaluation context}
=
\text{Evidence availability}
\times
\text{Evidence placement}\)
}
\\
\bottomrule
\end{tabularx}
\end{center}
\end{table*}

\begin{table}[htbp]
    \caption{\textbf{Gold-document positions.}
    Positions depend on the number of retained gold documents \(k\).}
    \centering
    \label{tab:gold_positions}
    \small
    \begin{tabular}{c c c c c c}
        \toprule
        \textbf{\(k\)} &
        \textbf{Front} &
        \textbf{Middle} &
        \textbf{End} \\
        \midrule
        1 & 1  & 10   & 19 \\
        2 & 1--2 & 10--11 & 19--20 \\
        3 & 1--3 & 9--11  & 18--20 \\
        4 & 1--4 & 9--12  & 17--20 \\
        \bottomrule
    \end{tabular}
\end{table}

\subsection{Evaluation Details}
\label{app:evaluation-details}

\paragraph{Answer normalization and scoring.} Predictions and references are normalized using Unicode NFKC normalization, lowercasing, removal of ASCII punctuation and the English articles
\emph{a}, \emph{an}, and \emph{the}, and whitespace normalization. The reference set contains the gold answer and its stored aliases. Exact Match (EM) is one if the normalized prediction is nonempty and matches any normalized reference. Token-level F1 is computed over whitespace-separated normalized tokens as
\[
    \mathrm{F1}(p,a)
    =
    \frac{2c(p,a)}{|p|+|a|},
\]
where \(c(p,a)\) is the multiset token overlap and \(|p|\) and \(|a|\) denote the respective token counts. We report the maximum score over all permissible references. For \texttt{yes}, \texttt{no}, and \texttt{noanswer}, pairwise F1 reduces to exact agreement. Empty or failed predictions receive zero, while missing requests are excluded.

\subsection{Reasoning Trap Construction}
\label{app:reasoning-traps}

\paragraph{Target-only evidence gap.} We construct the Reasoning Trap experiment from two-hop Bridge questions in the final \fondue\ release whose original reasoning path is preserved. Questions whose question or reasoning chain was modified through semantic repair are excluded. Each context uses the partial evidence gap \(G=\{f_1\}\), corresponding to a Target-only condition: the bridge document identifying the target entity is removed, while the true target document and its terminal evidence remain available. The model must therefore recover the missing bridge parametrically while distinguishing the true target from plausible alternatives.

\paragraph{Candidate construction.} Competitors must occupy the same functional position as the true target. They must be reachable through the same upstream relation, support the same terminal relation, and share at least one concrete \texttt{instance\_of} value with the true target. For example, for
\[
\textit{film} \rightarrow \textit{director} \rightarrow
\textit{country of citizenship},
\]
a competitor must likewise occur as the director of another work and have a supported citizenship value. Remaining positions are filled with \emph{bridge-like} documents whose entities match the type of the original bridge entity but do not satisfy the target-candidate information.

\paragraph{Evidence and leakage constraints.} Candidates are retrieved from the local knowledge graph and, where necessary, supplemented through targeted Wikidata queries. Every selected entity must have an article in the October 2017 English Wikipedia snapshot, and a competitor is retained only if its terminal value or a validated alias is supported by the article. Entities overlapping with the original reasoning path are excluded, and inserted articles may not mention original path entities. Bridge-like documents additionally may not mention selected competitors. Duplicate documents are removed, and all checks are applied to the complete retained article. These mention constraints use lexical matching over entity names and aliases.

\paragraph{Competition conditions.} For every question, we construct contexts with
\[
k \in \{0,1,4,9,14,19\}
\]
competitors. Each context contains exactly 20 documents and at most 30,000 \texttt{o200k\_base} tokens. The true target is fixed at rank~1, competitors occupy ranks \(2,\ldots,k+1\), and the remaining positions contain bridge-like documents:
\[
[\text{gold target},
 \underbrace{\text{competitors}}_{k},
 \underbrace{\text{bridge-like documents}}_{19-k}].
\]
Competitors therefore replace distractors rather than increasing context size. Candidate order is deterministic, and larger values of \(k\) extend the competitor prefix used at smaller values.

For \(k>0\), we construct three conditions.
\begin{description}
    \item[\textsc{Same-Gold}.]
    All competitors support the gold terminal value.

    \item[\textsc{Same-Other}.]
    All competitors support the same non-gold terminal value. The value is selected from an answer group that supports the required competition level.

    \item[\textsc{Random}.]
    Competitors are sampled from the full eligible pool without constraining their terminal values to agree.
\end{description}
Thus, \textsc{Same-Gold} and \textsc{Same-Other} control the semantic direction of competing evidence, whereas \textsc{Random} varies candidate competition without enforcing a common answer.

\paragraph{Coverage.}
Of the 5,138 Bridge questions in the release, 2,594 preserve their original reasoning path and enter candidate construction. After type, bridge-like document, and competitor availability checks, 2,207 questions retain a valid \(k=0\) control. The paired cohorts contain 1,527, 1,230, 1,060, 918, and 967 questions for \(k=1,4,9,14,19\), respectively. Analyses requiring complete coverage across all positive competition levels use the intersection of these cohorts (\(n=871\)). Candidate-competition experiments are reported for the subset of model configurations for which complete runs across all competition conditions are available.

\section{Additional Results}
\label{app:additional_results}

\subsection{Position-Matched Comparisons}
\label{app:position-matching}
\begin{figure}[t]
    \centering
    \includegraphics[height=0.35\textheight]{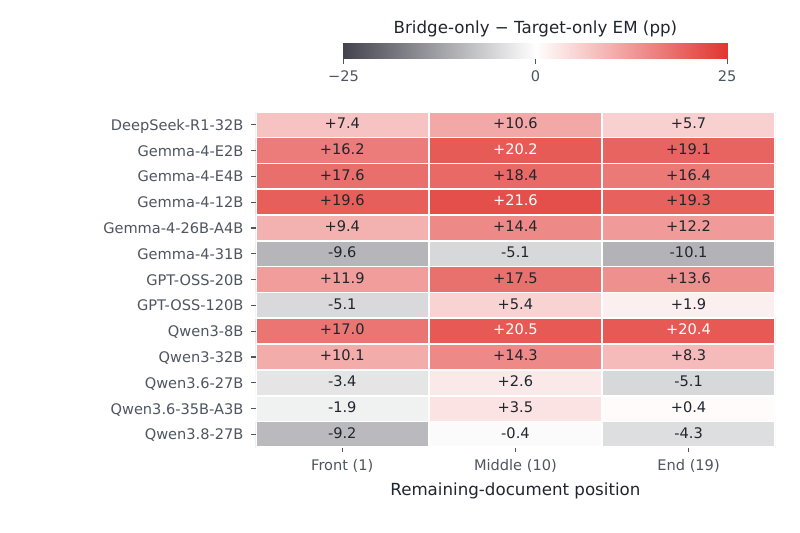}
    \caption{
    \textbf{Position-matched role comparison.}
    Difference in Exact Match between \tmpbridgeonly and \tmptargetonly when the retained gold document is placed at the same context position in both conditions. Positive values indicate higher performance with \tmpbridgeonly evidence.
    }
    \label{fig:position-matching}
\end{figure}

To control for document position, we compare \tmpbridgeonly and \tmptargetonly while placing the retained gold document at the same position in both conditions. We repeat this comparison with the retained document at the front, middle, and end of the context. \Cref{fig:position-matching} reports the resulting difference $\mathrm{EM}_{\text{Bridge-only}}-\mathrm{EM}_{\text{Target-only}}$, such that positive values indicate an advantage for bridge evidence.

The predominant role asymmetry persists across document positions. In particular, several smaller models retain a clear \tmpbridgeonly advantage at the front, middle, and end of the context. The magnitude varies with position, and some larger models remain close to zero or favor \tmptargetonly evidence, indicating that position can modulate but does not generally account for the observed role effect.

\subsection{Additional Candidate Competition Results}
\label{app:candidate-competition-results}

\begin{figure}[t]
    \centering
    \includegraphics[height=0.35\textheight]
        {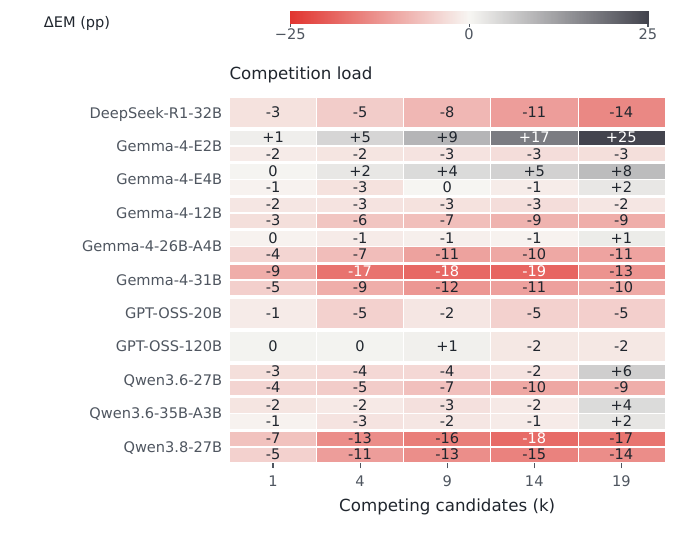}
    \caption{
    \textbf{Candidate competition across competition levels.}
    Change in Exact Match relative to the $k=0$ condition as the number
    of random type-compatible competitors increases.
    }
    \label{fig:app-competition-load}
\end{figure}

We provide additional model-level results for the candidate competition experiment from \Cref{sec:candidate_competition}. \Cref{fig:app-competition-load} shows the change in Exact Match across increasing competition levels for the random-candidate condition. The effect is heterogeneous across model configurations, but additional competitors reduce performance for many configurations, with larger reductions often appearing at higher competition levels.

\begin{figure}[t]
    \centering
    \includegraphics[height=0.04\textheight]{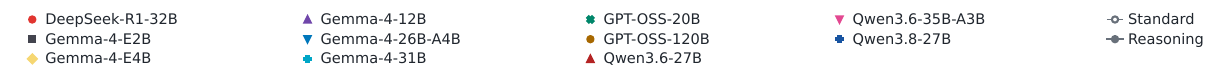}
    \includegraphics[height=0.21\textheight,keepaspectratio]{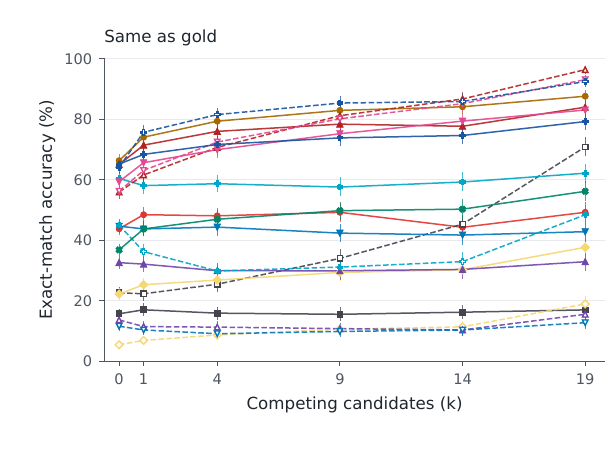}
    \hspace{0.015\textwidth}
    \includegraphics[height=0.21\textheight,keepaspectratio]{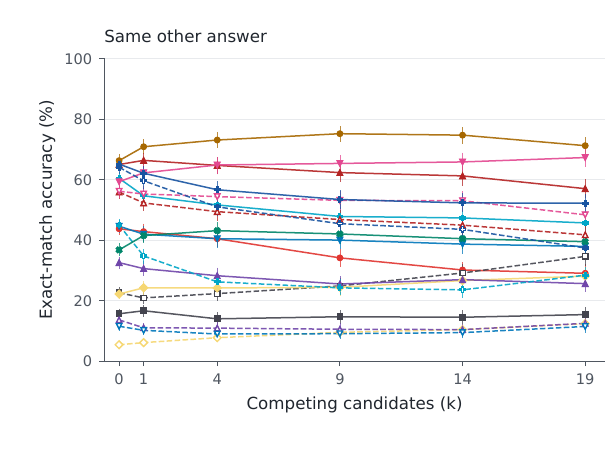}
    \caption{
    \textbf{Candidate-consistency effects across competition levels.}
    Exact Match with increasing numbers of competing candidates.
    Left: Same-Gold candidates support the correct terminal value.
    Right: Same-Other candidates consistently support an alternative value.
    }
    \label{fig:app-candidate-steering}
\end{figure}

\Cref{fig:app-candidate-steering} shows the complete trajectories for the \emph{Same-Gold} and \emph{Same-Other} conditions. \emph{Same-Gold} competitors generally shift performance toward the correct answer as competition increases, whereas \emph{Same-Other} competitors tend to shift performance in the opposite direction. The magnitude of both effects varies substantially across models and inference modes.

\subsection{Extended Benchmark Results}
\label{app:external-dataset-results}

We use the validated external benchmark variants to examine if effects observed on \fondue\ transfer to independently constructed multi-hop QA benchmarks. We evaluate \hotpotext, \twowikiext, and \mintqaext\ using the same available model configurations, scoring procedure, and evidence gap definitions wherever available. The resulting role analysis contains up to 5,123 \fondue\ questions, 2,252 \hotpotext\ questions, 3,545 \twowikiext\ questions, and 2,184 \mintqaext\ questions per model. For \hotpot, we additionally report the corresponding standard benchmark setting as a reference. We exclude \musiqueext\ from the main cross-benchmark analysis because only 531 instances remain after applying our validation pipeline. Where applicable, we also include results on standard \hotpot\ as a reference.

\begin{figure}[t]
    \centering
    \includegraphics[height=0.04\textheight]{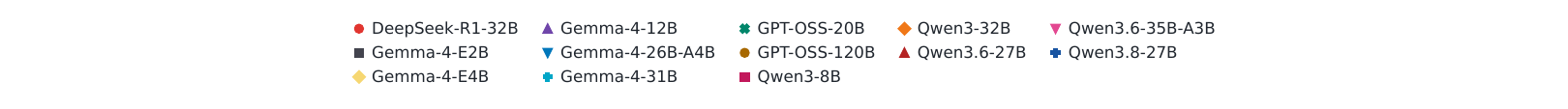}
    \includegraphics[height=0.18\textheight]{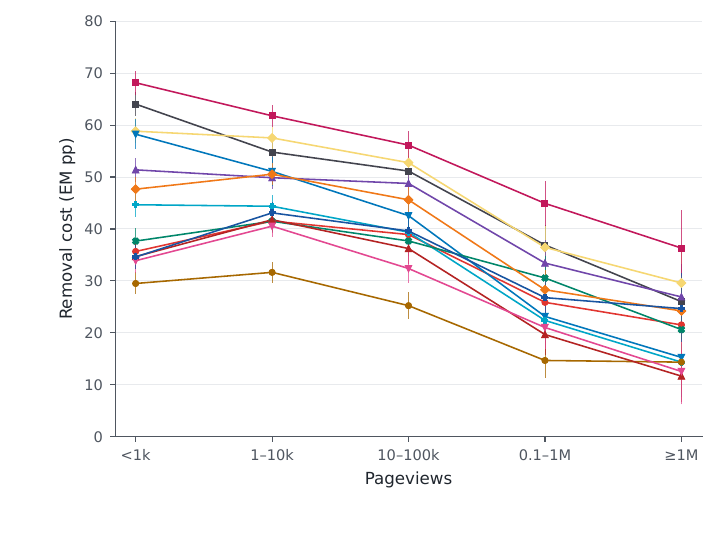}
    \hspace{0.015\textwidth}
    \includegraphics[height=0.18\textheight]{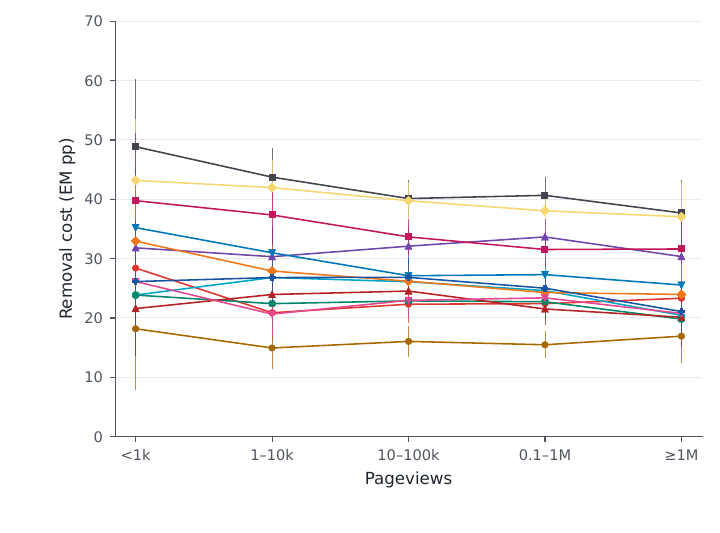}
    \hspace{0.015\textwidth}
    \includegraphics[height=0.18\textheight]{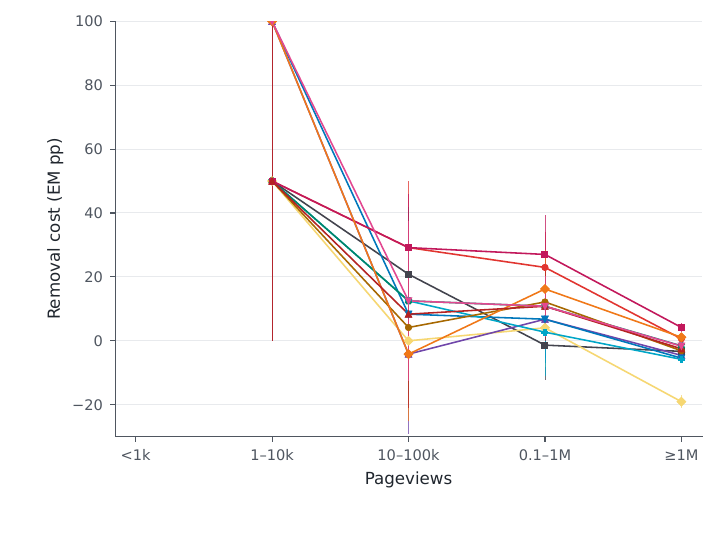}
    \caption{
    \textbf{Entity popularity and parametric compensation on the extended
    benchmarks.}
   Left: \twowikiext, Middle: \hotpotext, Right: \mintqaext}
    \label{fig:ext-popularity}
\end{figure}

\paragraph{Popularity effects generalize.}
We first examine whether the relationship between entity popularity and parametric compensation transfers beyond \fondue. As in the main analysis, we group all questions by the popularity of the entity associated with the removed fact and measure the resulting removal cost. The same general trend is visible on the external benchmarks, as missing evidence associated with more popular entities is typically less costly to remove, indicating that parametric compensation becomes easier as the relevant entity becomes more popular. However, the magnitude differs across datasets and models as can be seen in \Cref{fig:ext-popularity}.

\begin{figure}[t]
    \centering
    \includegraphics[width=0.30\linewidth]
        {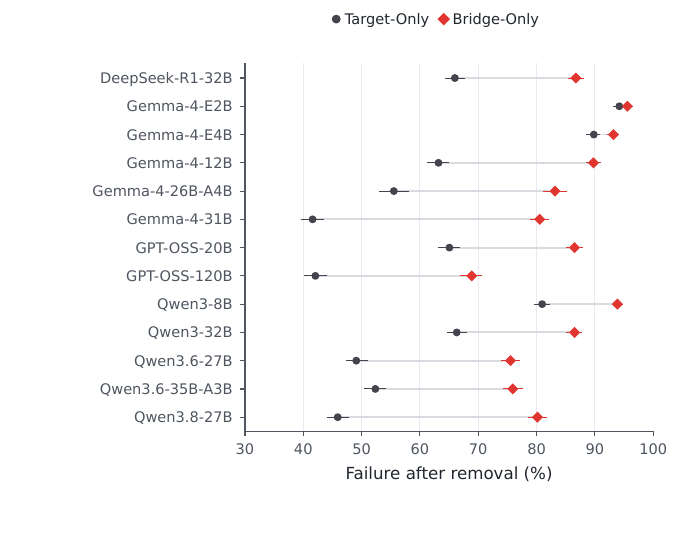}
    \hfill
    \includegraphics[width=0.30\linewidth]
        {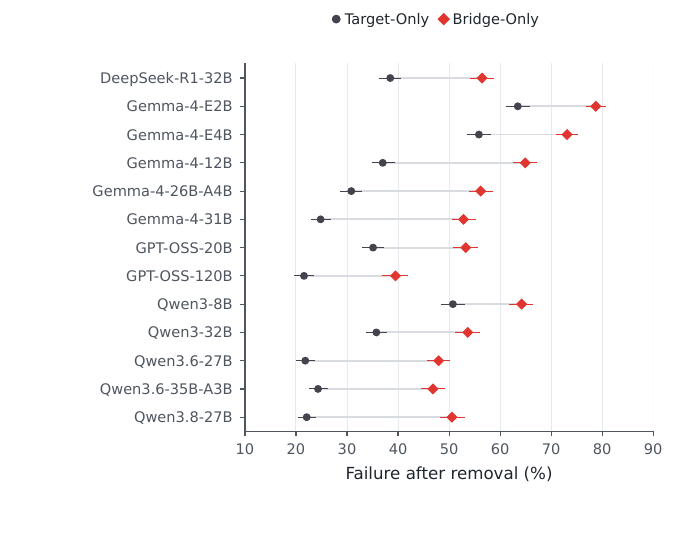}
    \hfill
    \includegraphics[width=0.30\linewidth]
        {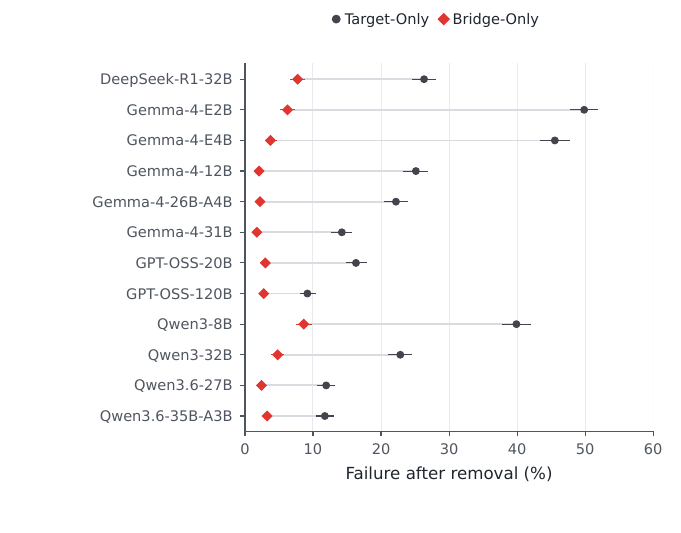}
    \caption{
    \textbf{Role-specific Evidence Gaps on the external benchmarks.}
    Failure after removing bridge or target evidence, conditioned on
    full-context success. \twowikiext\ (left) and \hotpotext\ (middle) predominantly favor Target-only evidence, whereas \mintqaext\ (right) strongly favors Bridge-only evidence.
    }
    \label{fig:ext-structural-role}
\end{figure}

\paragraph{Evidence-role effects are benchmark-dependent.}
As shown in \Cref{fig:ext-structural-role}, the structural role effect transfers less uniformly. On \fondue, Bridge-only evidence outperforms Target-only evidence for most models, while \mintqaext\ shows the same direction and generally an even larger Bridge-only advantage. In contrast, \twowikiext\ and \hotpotext\ predominantly favor Target-only evidence. The corresponding standard \hotpot\ setting shows the same tendency, but with a weaker asymmetry.

\begin{figure}[t]
    \centering
    \includegraphics[width=0.48\linewidth]
        {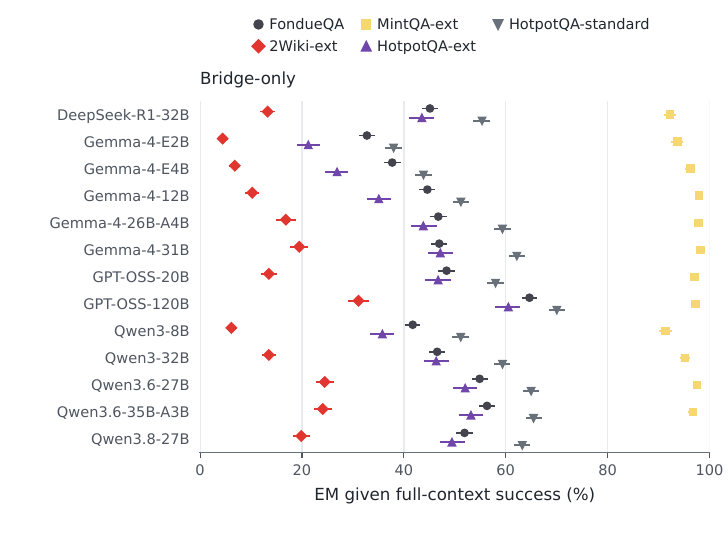}
    \hfill
    \includegraphics[width=0.48\linewidth]
        {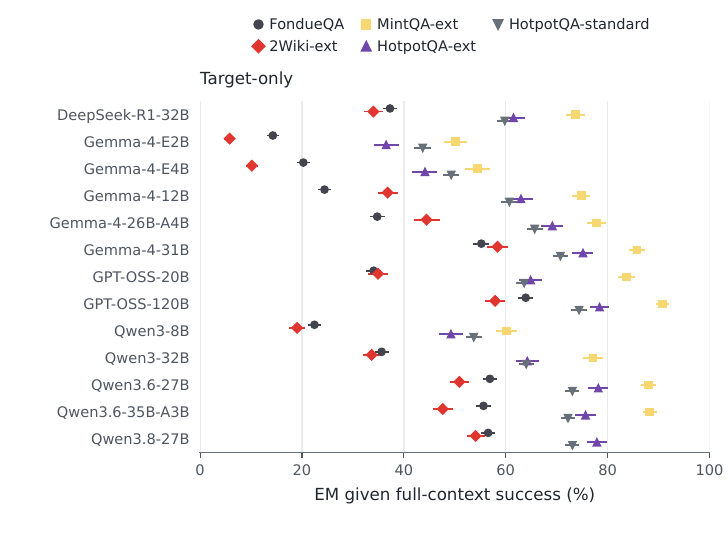}
    \caption{
    \textbf{Decomposition of the cross-benchmark role effect.}
    Partial-context EM conditioned on full-context success for Bridge-only and Target-only evidence. Differences across benchmarks are particularly pronounced in the Bridge-only condition.
    }
    \label{fig:ext-role-decomposition}
\end{figure}

To identify where these differences arise, we separately compare Bridge-only and Target-only accuracy across datasets (\Cref{fig:ext-role-decomposition}). The largest discrepancies appear in the Bridge-only condition. Models retain substantially more of their full-context successes from bridge evidence on \fondue\ and especially \mintqaext, whereas Bridge-only accuracy is markedly lower on \twowikiext.

\begin{figure}[t]
    \centering
    \includegraphics[width=0.48\linewidth]
        {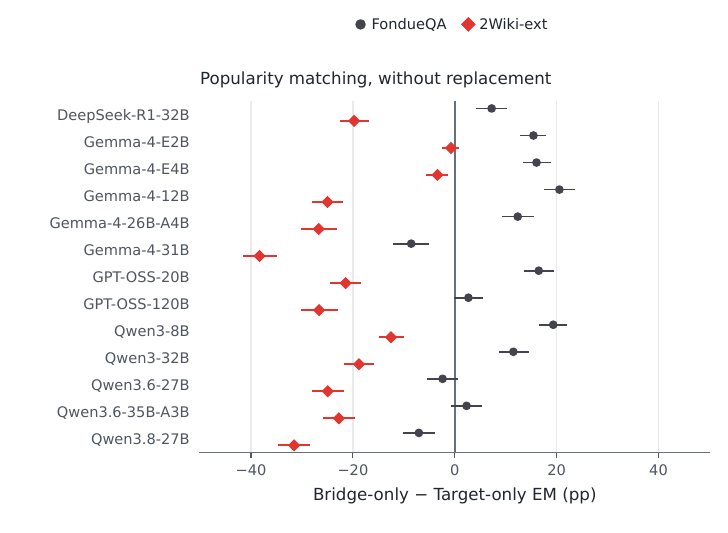}
    \hfill
    \includegraphics[width=0.48\linewidth]
        {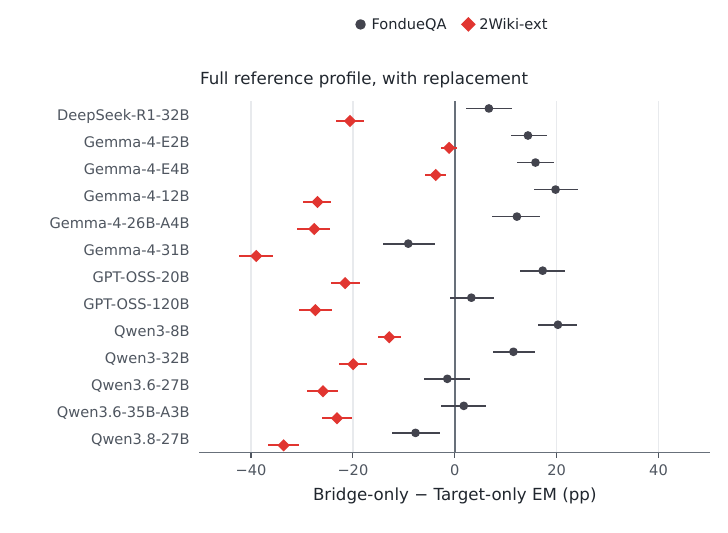}
    \caption{
    \textbf{Role effects after controlling for entity popularity.} Bridge-only minus Target-only EM after one-to-one popularity matching without replacement (left) and reweighting \fondue\ toward the full \twowikiext\ popularity profile (right). The cross-benchmark reversal persists under both controls.
    }
    \label{fig:ext-popularity-controls}
\end{figure}

\paragraph{Popularity does not explain the \twowikiext\ reversal.} A possible explanation for the different role effect is that \fondue\ and \twowikiext\ contain entities with different levels of popularity. We therefore align their bridge- and target-entity popularity distributions (\Cref{fig:ext-popularity-controls}).

First, one-to-one matching without replacement yields 1,892 questions from each benchmark. The resulting bridge medians are 3,479 pageviews for \fondue\ and 3,466 for \twowikiext, while the target medians are 10,534 and 10,412, respectively. Despite this alignment, the role-effect reversal remains.

We additionally reweight \fondue\ toward the complete \twowikiext\ popularity profile. The resulting bridge medians are 2,659 and 2,630 pageviews, respectively, and the target medians are 7,569 and 7,428. The weighted \fondue\ sample contains 1,289 unique questions with an effective sample size of approximately 846, while the reference \twowikiext\ profile contains 2,124 questions. The qualitative difference again remains. Differences in role-entity popularity alone therefore do not explain the \twowikiext\ reversal.

\begin{figure}[t]
    \centering
    \includegraphics[width=0.60\linewidth]
        {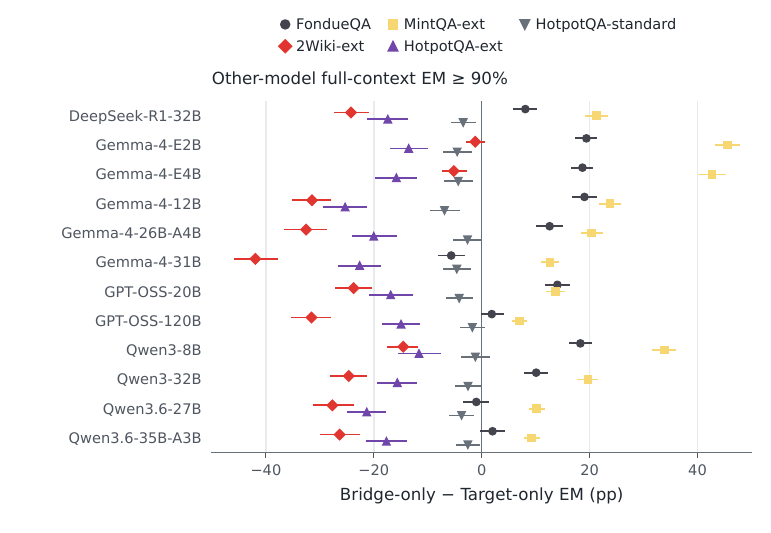}
    \caption{
    \textbf{Role effects on questions reliably solved by other models.} Bridge-only minus Target-only EM for questions on which the remaining evaluated models achieve at least 90\% mean full-context EM.
    }
    \label{fig:ext-difficulty}
\end{figure}

\paragraph{The reversal persists on easier questions.} We next test whether differences in overall question difficulty account for the role effect. For each evaluated model, we retain questions for which the other models achieve at least 90\% mean full-context EM (\Cref{fig:ext-difficulty}), excluding the evaluated model from the difficulty estimate. This leaves 2,921--3,260 \fondue, 1,035--1,093 \hotpotext, 836--928 \twowikiext, and 1,747--1,791 \mintqaext\ questions per model. For the standard \hotpot\ reference, 2,601--2,740 questions remain.

The benchmark-specific pattern remains under this restriction. \twowikiext\ and \hotpotext\ continue to favor Target-only evidence for most models, whereas \mintqaext\ remains strongly Bridge-only-favoring. The standard \hotpot\ setting shows a weaker Target-only tendency. \fondue\ also remains Bridge-only-favoring for most models, although several effects move closer to zero. Hence, restricting the analysis to questions that are reliably answered from
complete evidence does not remove the benchmark-specific role differences.

\paragraph{Discussion.}
The external results show different degrees of generalization for the two determinants studied in \Cref{sec:eval-drivers}. Entity popularity transfers comparatively well, as across benchmarks, missing information about more popular entities is generally easier to compensate for parametrically. The relative utility of bridge and target evidence, however, is benchmark-dependent. Bridge evidence can be more useful than the answer-bearing document, as observed on \fondue\ and \mintqaext. In contrast, \twowikiext\ and \hotpotext\ predominantly favor Target-only evidence, while the standard \hotpot\ setting shows a weaker Target-only tendency.

Neither aligning entity popularity nor restricting the analysis to easier questions removes the difference between \fondue\ and \twowikiext. The datasets nevertheless differ substantially in their context construction (\Cref{tab:ext-context-comparison}).

\begin{table}[t]
    \centering
    \small
    \caption{
    \textbf{Context characteristics for the role analysis.}
    Token counts denote median inference-time context lengths.
    }
    \label{tab:ext-context-comparison}
    \begin{tabular}{lrr}
        \toprule
        \textbf{Dataset}
        & \textbf{Documents}
        & \textbf{Tokens} \\
        \midrule
        \fondue          & 20 & 7,243 \\
        \hotpotext & 10 & 1,268 \\
        \hotpot      & 10 & 1,383 \\
        \twowikiext & 10 & 732 \\
        \mintqaext    & 10 & 952 \\
        \bottomrule
    \end{tabular}
\end{table}

Context length alone is unlikely to explain the benchmark-specific role differences. All external settings considered here use ten-document contexts and are substantially shorter than \fondue, yet their role effects point in different directions. A possible difference instead lies in the composition of the remaining context and in how many plausible entities or bindings remain after an evidence gap is introduced. This would be consistent with the candidate competition experiment (\Cref{sec:candidate_competition}), where plausible alternative targets change the usefulness of otherwise available evidence.

The external comparison does not manipulate these properties directly. Context composition and candidate ambiguity should therefore be interpreted as possible explanations rather than established causes of the cross-benchmark differences. Overall, the results suggest that parametric accessibility through entity popularity generalizes comparatively robustly, whereas evidence roles do not admit a universal ordering of importance. Their usefulness depends on the broader question-context configuration in which the missing reasoning step must be recovered.

\end{document}